\documentclass[letterpaper, 10 pt, conference]{ieeeconf}
\IEEEoverridecommandlockouts
\usepackage{amsmath,amsfonts}
\usepackage{array}
\usepackage{textcomp}
\usepackage{stfloats}
\usepackage{url}
\usepackage{verbatim}
\usepackage{graphicx}
\usepackage{cite}

\usepackage[utf8]{inputenc}
\usepackage{hyperref}

\usepackage{subcaption}
\usepackage{bm}
\usepackage{rotating}
\usepackage{multirow}
\usepackage{hhline}

\usepackage{soul}
\usepackage{afterpage}
\usepackage{color}
\usepackage{latexsym}
\usepackage{amssymb}
\usepackage{mathtools}
\usepackage{pdflscape}
\usepackage{enumerate}
\let\labelindent\relax
\usepackage[inline]{enumitem}
\usepackage{xspace} 
\usepackage{float}
\usepackage{booktabs}
\usepackage{diagbox}
\usepackage{tabu}
\usepackage{colortbl}

\newcounter{rowcntr}[table]
\renewcommand{\therowcntr}{\thetable.\arabic{rowcntr}}

\newcolumntype{N}{>{\refstepcounter{rowcntr}\therowcntr}c}

\AtBeginEnvironment{table}{\setcounter{rowcntr}{0}}

\newcommand\rowtag[2]{#1\def\@currentlabel{#1}\label{#2}}

\usepackage{pifont} 

\usepackage[dvipsnames]{xcolor}
\usepackage[normalem]{ulem}
\usepackage{makecell}
\usepackage{algorithm2e}
\usepackage{multicol}
\usepackage[capitalise]{cleveref}
\usepackage[percent]{overpic}

\usepackage{tikz}
\usetikzlibrary{arrows.meta,arrows,shapes,tikzmark,calc,backgrounds,decorations.pathreplacing,calligraphy,shapes.multipart}

\usepackage{cuted}
\usepackage{fancyhdr}

\usepackage[per-mode=fraction]{siunitx}
\usepackage{comment}
\usepackage[normalem]{ulem}
\usepackage{xspace}
\usepackage{xcolor}
\usepackage{lastpage}
\usepackage{fancyhdr}

\DeclareSIUnit\px{px}
\DeclareSIUnit\fps{fps}

\definecolor{OliveGreen}{RGB}{0,200,25}
\newcommand{\red}[1]{\textcolor{red}{#1}}

\newcommand{\darkgreen}[1]{\textcolor{OliveGreen}{#1}}

\newcommand{\ie}{i.\,e.\xspace}
\newcommand{\eg}{e.\,g.\xspace}

\newcommand{\armarVI}{\mbox{ARMAR-6}\xspace}

\newcommand{\ackJuBot}{This work has been supported by the Carl Zeiss Foundation through the JuBot project.}

\newif\iffinal

\newcommand{\enablefinalversion}{\finaltrue}

\newcommand{\replaced}[2]{%
	\iffinal%
	#2%
	\else%
	\red{\ifmmode\text{\sout{\ensuremath{#1}}}\else\sout{#1}\fi}\darkgreen{#2}%
	\fi%
}
\newcommand{\removed}[1]{%
	\iffinal%
	\else%
	\red{\ifmmode\text{\sout{\ensuremath{#1}}}\else\sout{#1}\fi}%
	\fi%
}

\definecolor{cadmiumyellow}{rgb}{1.0, 0.96, 0.0}
\definecolor{canaryyellow}{rgb}{1.0, 0.94, 0.0}
\definecolor{bananayellow}{rgb}{1.0, 0.88, 0.21}
\definecolor{yellow(process)}{rgb}{1.0, 0.94, 0.0}
\definecolor{applegreen}{rgb}{0.55, 0.71, 0.0}
\definecolor{capri}{rgb}{0.0, 0.75, 1.0}
\definecolor{corn}{rgb}{0.98, 0.93, 0.36}
\definecolor{cornflowerblue}{rgb}{0.39, 0.58, 0.93}
\definecolor{darkblue}{rgb}{0.0, 0.0, 0.55}
\definecolor{burntorange}{rgb}{0.8, 0.33, 0.0}
\definecolor{cadmiumred}{rgb}{0.89, 0.0, 0.13}
\definecolor{capri}{rgb}{0.0, 0.75, 1.0}
\definecolor{amber}{rgb}{1.0, 0.75, 0.0}
\definecolor{prune}{rgb}{0.44, 0.11, 0.11}
\definecolor{purplepizzazz}{rgb}{1.0, 0.31, 0.85}
\definecolor{amber_sae}{rgb}{1.0, 0.49, 0.0}
\definecolor{alizarin}{rgb}{0.82, 0.1, 0.26}
\definecolor{darkorange}{rgb}{1.0, 0.55, 0.0}
\definecolor{darkpowderblue}{rgb}{0.0, 0.2, 0.6}
\definecolor{darkpastelpurple}{rgb}{0.59, 0.44, 0.84}
\definecolor{darkpink}{rgb}{0.91, 0.33, 0.5}

\usepackage{tikz-imagelabels}

\DeclareCaptionLabelFormat{custom}
{%
}
\DeclareCaptionLabelSeparator{custom}{--}
\DeclareCaptionFormat{custom}
{%
}

\usepackage{mathrsfs}

\newcommand{\lf}{\mathcal{F}}

\newcommand{\ptop}{\mathsf{p2p}}
\newcommand{\ptol}{\mathsf{p2l}}
\newcommand{\ptoP}{\mathsf{p2P}}
\newcommand{\ptoc}{\mathsf{p2c}}
\newcommand{\ptoS}{\mathsf{p2S}}
\newcommand{\pose}{\mathsf{pose}}

\newcommand{\set}[1]{\{#1\}}

\newcommand{\scaleeq}[2]{\scalebox{#1}{$#2$}}

\newcommand{\mast}[2]{\scalebox{0.6}{$\mathsf{M}^{#1}_{#2}$}}
\newcommand{\slav}[1]{\scalebox{0.6}{$\mathsf{S}_{#1}$}}
\newcommand{\obj}[1]{$\mathsf{#1}$}

\newcommand{\master}{\scaleeq{0.95}{\mathsf{master}}}
\newcommand{\slave}{\scaleeq{0.95}{\mathsf{slave}}}
\newcommand{\task}[2]{\scalebox{0.95}{$\mathsf{#1}$ \raisebox{0.5pt}{\circled{#2}}}}
\newcommand{\taskabbr}[1]{\scalebox{0.95}{$\mathsf{#1}$}}

\DeclareMathOperator*{\argmin}{arg\,min}

\DeclareRobustCommand\sampleline[1]{
    \tikz\draw[#1] (0,0) (0,\the\dimexpr\fontdimen22\textfont2\relax) -- (1em,\the\dimexpr\fontdimen22\textfont2\relax);%
}

\DeclareRobustCommand\inlinearrow[1]{ 
    \hspace{-0.7em}
    \tikz{
        \draw[-{to[length=1.5mm]},
        #1,line width=1.1pt](0,0) (0,\the\dimexpr\fontdimen22\textfont2\relax) -- (1em,\the\dimexpr\fontdimen22\textfont2\relax);
    }
    \hspace{-0.7em}
}

\DeclareRobustCommand\circled[1]{\tikz[baseline=(char.base)]{
		\node[shape=circle,fill,inner sep=2pt,scale=0.3] (char) {\textcolor{white}{\Huge \textbf{#1}}};}}

\usepackage{amsmath}

\newenvironment{talign*}
 {\let\displaystyle\textstyle\csname align*\endcsname}
 {\endalign}

\newcolumntype{P}[1]{>{\centering\arraybackslash}p{#1}}
\newcolumntype{M}[1]{>{\centering\arraybackslash}m{#1}}

\newlength\imgwidth
\newlength\imgheight

\newsavebox{\subfigbox}

\enablefinalversion

\def\BibTeX{{\rm B\kern-.05em{\sc i\kern-.025em b}\kern-.08em
    T\kern-.1667em\lower.7ex\hbox{E}\kern-.125emX}}

\begin{document}

\thispagestyle{fancy}	
\fancyhead[RO,LE]{Accepted in IEEE International Conference on Robotics and Automation (ICRA) 2024}
\fancyhead{}

\title{\LARGE \bf
	Bi-KVIL: Keypoints-based Visual Imitation Learning \\of Bimanual Manipulation Tasks
	\thanks{\ackJuBot}
	\thanks{
		The authors are with the Institute for Anthropomatics and Robotics, Karlsruhe Institute of Technology, Karlsruhe, Germany. E-mails: \{jianfeng.gao, asfour\}@kit.edu}
	\thanks{This work has been submitted to the IEEE for possible publication. Copyright may be transferred without notice, after which this version may no longer be accessible. Accepted in IEEE International Conference on Robotics and Automation (ICRA) 2024.}
}

\author{
Jianfeng~Gao,
Xiaoshu Jin,
Franziska Krebs,  
No\'emie Jaquier, 
and~Tamim~Asfour
}

\makeatletter
\let\@oldmaketitle\@maketitle
\renewcommand{\@maketitle}{\@oldmaketitle
	\vspace{-11ex}
}
\makeatother
\maketitle

\begin{strip}
\captionsetup[sub]{labelformat=parens}
\begingroup
    \captionsetup{type=figure}
    \begin{subfigure}[t]{0.20\textwidth}
        \centering
        \resizebox{!}{48mm}{\begin{tikzpicture}
    \node (image_1) at (0,0) {
         \includegraphics[height=28mm]{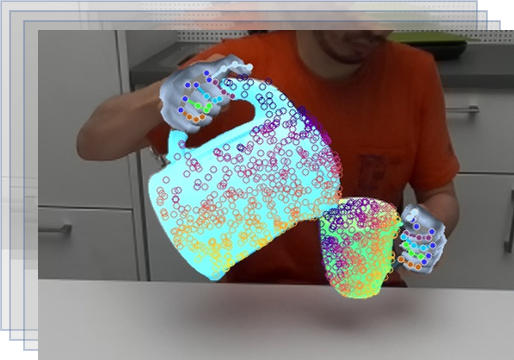}
    };
    
    \node (image_2) at (0,-2.9) {
         \includegraphics[height=28mm]{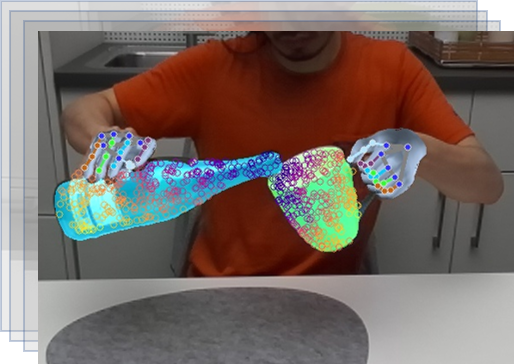}
    };

    \node[rotate=90] at ($(image_1.west) + (-0.3, 0)$) {Pour Water Task};
    \node[rotate=90] at ($(image_2.west) + (-0.3, 0)$) {Pour Beer Task};

\end{tikzpicture}}
        \caption{\small{Human demonstrations.}\label{subfig:teaser_demo}}
    \end{subfigure}
    \begin{subfigure}[t]{0.12\textwidth}
        \centering
        \resizebox{!}{48mm}{
        \begin{tikzpicture}
            \node (image) at (0,0) {
                \resizebox{!}{28mm}{
                    \begin{tikzpicture}
    \node[draw, fill=blue!15, ellipse, minimum width=8mm, minimum height=7mm] (m0) at (0, 0) {\obj{vc}};       
    \node[draw, fill=green!15, ellipse, minimum width=8mm, minimum height=7mm] (m1_0) at (0, -1.2) {\obj{c}};       
    \node[draw, fill=red!15, ellipse, minimum width=8mm, minimum height=7mm] (m2_0) at (0, -3) {\obj{lh}};       
    \node[draw, fill=green!15, ellipse, minimum width=8mm, minimum height=7mm] (m2_1) at ( 1.0, -2) {\obj{k}};
    \node[draw, fill=red!15, ellipse, minimum width=8mm, minimum height=7mm] (s0) at ( 1.0, -3) {\obj{rh}};
    
    \draw[-latex, ForestGreen] (m1_0) -- (m0) node[pos=0.5, right] {\scalebox{0.95}{\obj{pose}}};
    \draw[-latex, orange] (m2_0) -- (m1_0) node[pos=0.13, right=1mm] {\scalebox{0.95}{\obj{g}}};
    \draw[-latex, blue] (m2_1) -- (m1_0) node[pos=1.0, right=1mm] {\scalebox{0.95}{\obj{\ptop, \ptoc}}};
    \draw[-latex, orange] (s0) -- (m2_1) node[pos=0.5, right=1mm] {\scalebox{0.95}{\obj{g}}};
    
\end{tikzpicture}
                }
            };
            \node (image) at (0,-2.9) {
                \resizebox{!}{28mm}{
                    \begin{tikzpicture}
    \node[draw, fill=blue!15, ellipse, minimum width=8mm, minimum height=7mm] (m0) at (0, 0) {\obj{vc}};       
    \node[draw, fill=green!15, ellipse, minimum width=8mm, minimum height=7mm] (m1_0) at (0, -1.2) {\obj{c}};       
    \node[draw, fill=red!15, ellipse, minimum width=8mm, minimum height=7mm] (m2_0) at (0.0, -3) {\obj{lh}};       
    \node[draw, fill=green!15, ellipse, minimum width=8mm, minimum height=7mm] (m2_1) at ( 1.0, -2) {\obj{b}};
    \node[draw, fill=red!15, ellipse, minimum width=8mm, minimum height=7mm] (s0) at ( 1.0, -3) {\obj{rh}};
    
    \draw[-latex, ForestGreen] (m1_0) -- (m0) node[pos=0.5, right] {\scalebox{0.95}{\obj{pose}}};
    \draw[-latex, ForestGreen] (m1_0) -- (m0) node[pos=0.5, right] {\scalebox{0.95}{\obj{pose}}};
    \draw[-latex, blue] (m2_1) -- (m1_0) node[pos=1.0, right=1mm] {\scalebox{0.95}{\obj{\ptop, \ptoc}}};
    \draw[-latex, blue] (m2_1) -- (m1_0) node[pos=1.0, right=1mm] {\scalebox{0.95}{\obj{\ptop, \ptoc}}};
    \draw[-latex, darkorange] (s0) -- (m2_1) node[pos=0.5, right=1mm] {\scalebox{0.95}{\obj{g}}};
    \draw[-latex, darkorange] (m2_0) -- (m1_0) node[pos=0.13, right=1mm] {\scalebox{0.95}{\obj{g}}};
    \draw[-latex, darkorange] (s0) -- (m2_1) node[pos=0.5, right=1mm] {\scalebox{0.95}{\obj{g}}};
    \draw[-latex, darkorange] (m2_0) -- (m1_0) node[pos=0.13, right=1mm] {\scalebox{0.95}{\obj{g}}};
  
\end{tikzpicture}
                }
            };
        \end{tikzpicture}
        }
        \caption{\small{HMSR.}\label{subfig:teaser_msr}}
    \end{subfigure}
    \begin{subfigure}[t]{0.25\textwidth}
        \centering
        \resizebox{!}{48mm}{\begin{tikzpicture}
    \node (image) at (0,0) {
         \includegraphics[height=28mm]{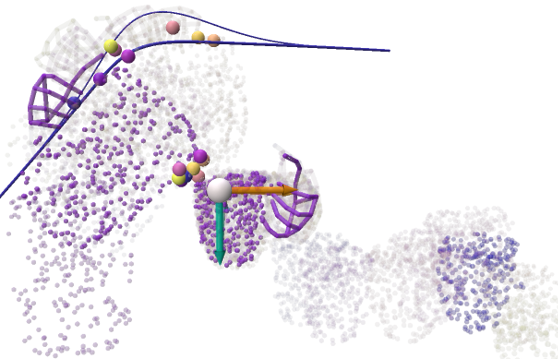}
    };
    \node (image) at (0,-2.9) {
         \includegraphics[height=28mm]{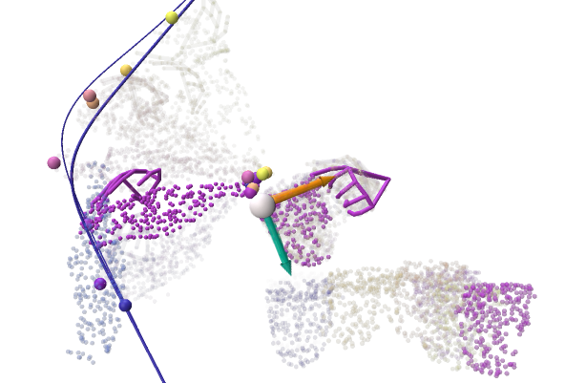}
    };


    \node at (-0.80, -0.3) {\scalebox{1.0}{$\lf$}};
    \node at (-0.25,  0.3) {\scalebox{0.8}{\textcolor{blue}{$\ptop$}}};
    \node at ( 0.40,  0.1) {\scalebox{0.8}{\obj{lh}}};
    \node at (-2.20,  0.8) {\scalebox{0.8}{\obj{rh}}};
    \node at (-0.05,  0.8) {\scalebox{0.8}{\textcolor{ForestGreen}{$\ptoc$}}};
    \node at (-1.60,  1.2) {\scalebox{0.8}{density}};
    
    \node at (-0.35, -0.8) {\scalebox{0.8}{\obj{c}}};
    \node at ( 1.00, -1.0) {\scalebox{0.8}{\obj{vc}}};
    \node at (-2.20, -0.2) {\scalebox{0.8}{\obj{k}}};
    \node at (-2.20, -1.0) {\scalebox{0.8}{\obj{vk}}};

    \node at (-0.25, -3.3) {\scalebox{1.0}{$\lf$}};
    \node at (-0.00, -2.6) {\scalebox{0.8}{\textcolor{blue}{$\ptop$}}};
    \node at ( 0.90, -3.0) {\scalebox{0.8}{\obj{lh}}};
    \node at (-1.00, -2.6) {\scalebox{0.8}{\obj{rh}}};
    \node at (-0.70, -2.0) {\scalebox{0.8}{\textcolor{ForestGreen}{$\ptoc$}}};
    \node at (-1.90, -2.1) {\scalebox{0.8}{density}};
    
    \node at ( 0.40, -3.3) {\scalebox{0.8}{\obj{c}}};
    \node at ( 1.00, -3.9) {\scalebox{0.8}{\obj{vc}}};
    \node at (-1.80, -3.1) {\scalebox{0.8}{\obj{b}}};
    \node at (-1.80, -3.9) {\scalebox{0.8}{\obj{vb}}};
    
    \node at (-0.8, -1.20) {\scalebox{0.8}{movement primitives}};
    \draw[dashed] (2.7, 1.3) -- (2.7, -4.3);

    \node[rectangle, rounded corners, minimum width=18mm, minimum height=13mm, fill=cyan!10, draw=black!20] at (1.55, 0.75) {};
            
    \coordinate (c)  at (0.6, 1.2);
    \coordinate (k)  at (0.6, 0.9);
    \coordinate (vc) at (0.6, 0.6);
    \coordinate (vk) at (0.6, 0.3);
    \node[right, fill=none, text=black] at (vc.east) {\scalebox{0.7}{\obj{vc}: virtual cup}};
    \node[right, fill=none, text=black] at (vk.east) {\scalebox{0.7}{\obj{vk}: virtual kettle}};
    \node[right, fill=none, text=black] at  (c.east) {\scalebox{0.7}{\obj{c}: cup}};
    \node[right, fill=none, text=black] at  (k.east) {\scalebox{0.7}{\obj{k}: \ kettle}};

    \node[rectangle, rounded corners, minimum width=18mm, minimum height=13mm, fill=cyan!10, draw=black!20] at (1.55, -2.15) {};
    \coordinate (b)  at (0.6, -1.7);
    \coordinate (vb) at (0.6, -2.0);
    \coordinate (lh) at (0.6, -2.3);
    \coordinate (rh) at (0.6, -2.6);
    
    \node[right, fill=none, text=black] at ( b.east)  {\scalebox{0.7}{\obj{b}: \ beer}};
    \node[right, fill=none, text=black] at (vb.east) {\scalebox{0.7}{\obj{vb}: virtual beer}};
    \node[right, fill=none, text=black] at (lh.east) {\scalebox{0.7}{\obj{lh}: left hand}};
    \node[right, fill=none, text=black] at (rh.east) {\scalebox{0.7}{\obj{rh}: right hand}};
    
\end{tikzpicture}}
        \caption{\small{Geometric task representation.}\label{subfig:teaser_constr}} 
    \end{subfigure}
    \begin{subfigure}[t]{0.4\textwidth}
        \centering
        \resizebox{!}{48mm}{\begin{tikzpicture}
    \node (image) at (0, 0) {
        \includegraphics[height=24mm, trim=20 0 20 0, clip]{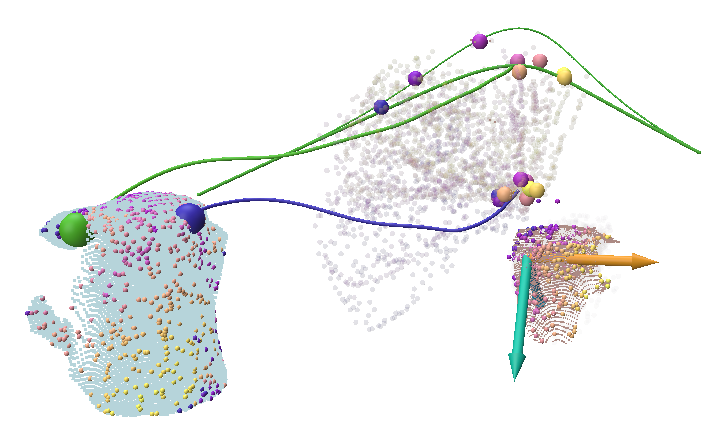}
    };
    \node (image) at (4.5, 0) {
         \includegraphics[height=24mm]{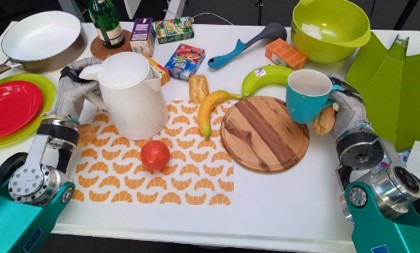}
    };
    \node (image) at (0,-2.9) {
         \includegraphics[height=32mm]{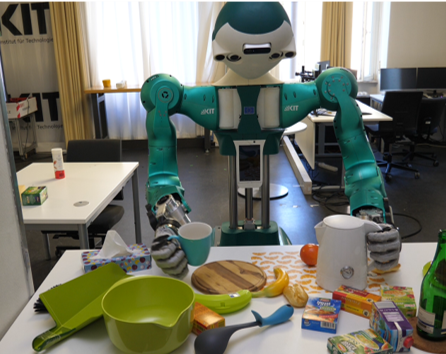}
    };
    \node (image) at (4.5,-2.9) {
         \includegraphics[height=32mm]{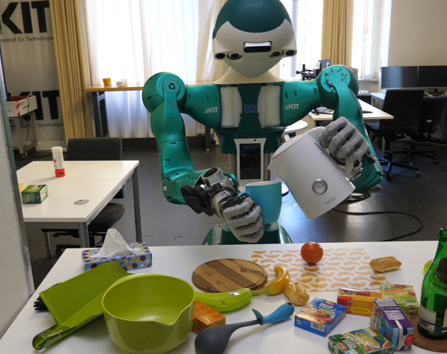}
    };
    
    \node at ( 0.6,  -0.3) {\scalebox{1.0}{$\lf$}};
    \node at ( 1.4,   0.1) {\scalebox{0.8}{\textcolor{blue}{$\ptop$}}};
    \node at (-0.1,  -0.3) {\scalebox{0.8}{\textcolor{blue}{MP$_{\ptop}$}}};
    \node at (-0.9,   0.5) {\scalebox{0.8}{\textcolor{ForestGreen}{MP$_{\ptoc}$}}};
    \node at ( 1.7,   0.8) {\scalebox{0.8}{\textcolor{ForestGreen}{$\ptoc$}}};
    
    \node at ( 1.30, -0.7) {\scalebox{0.8}{\obj{c}}};
    \node at (-1.80, -0.9) {\scalebox{0.8}{\obj{k}}};
\end{tikzpicture}}
        \caption{\small{Reproduction.}}
        \label{subfig:teaser_exe}
    \end{subfigure}
    \caption{Overview of Bi-KVIL. 
        \subref{subfig:teaser_demo} Human demonstration videos of the pouring tasks are collected with different styles and pose variations of categorical objects. 
        \subref{subfig:teaser_msr} For each task, we abstract the hand/object relationships into a symbolic \emph{Hybrid Master- Slave Relationship} (HMSR) with
        \subref{subfig:teaser_constr} sub-symbolic geometric constraints for each \master{}-\slave{} pair to model motion styles.
        \subref{subfig:teaser_exe} The learned tasks are then reproduced with category-level generalization in cluttered scenes by \armarVI.}
    \label{fig:overview} 
    \vspace{-2ex}
\endgroup
\end{strip}

\begin{abstract}
Visual imitation learning has achieved impressive progress in learning unimanual manipulation tasks from a small set of visual observations, thanks to the latest advances in computer vision.
However, learning bimanual coordination strategies and complex object relations from bimanual visual demonstrations, as well as generalizing them to categorical objects in novel cluttered scenes remain unsolved challenges. 
In this paper, we extend our previous work on keypoints-based visual imitation learning (\mbox{K-VIL})~\cite{gao_kvil_2023} to bimanual manipulation tasks. The proposed Bi-KVIL jointly extracts so-called \emph{Hybrid Master-Slave Relationships} (HMSR) among objects and hands, bimanual coordination strategies, and sub-symbolic task representations. 
Our bimanual task representation is object-centric, embodiment-independent, and viewpoint-invariant, thus generalizing well to categorical objects in novel scenes. 
We evaluate our approach in various real-world applications, showcasing its ability to learn fine-grained bimanual manipulation tasks from a small number of human demonstration videos.
Videos and source code are available at \href{https://sites.google.com/view/bi-kvil}{https://sites.google.com/view/bi-kvil}.
\end{abstract}


\vspace{-0.2cm}
\section{Introduction}
\label{sec:introduction}

Bimanual manipulation is key to human everyday activities while being significantly more complex than the simple sum of two unimanual tasks.
Similarly to the unimanual case, bimanual 
manipulation tasks are usually characterized by invariant task features over several demonstrations~\cite{muhlig_automatic_2009,muhlig_task-level_2009}. For instance, pouring tasks are characterized via the alignment of the mouth of the container to the rim of the cup (see \cref{subfig:teaser_demo}).
In~\cite{gao_kvil_2023}, we introduced the Keypoint-based Visual Imitation Learning (\mbox{K-VIL}) framework that leverages this principle to automatically extract sparse, object-centric, and embodiment-independent task representations from few
human demonstration videos.
However, K-VIL is limited to unimanual tasks with a single \master{}-\slave{} pair and a static \master{} object. 
Instead, bimanual tasks often involve more than two objects, as well as more complex \master{}-\slave{} relationships since each \master{} object may itself be a motion-salient \slave{} object paired to another \master{} object. 

In this paper, we build on our previous work~\cite{gao_kvil_2023} and propose Bi-KVIL, an approach for learning bimanual task representations that capture all relevant temporal and spatial constraints between the hands/objects
(see \cref{subfig:teaser_msr,subfig:teaser_constr}).
To this end, it is important to understand their roles and relationships in the demonstrated bimanual tasks.
Early works focused on hand/arm relationships but overlooked the role of objects.
For example, \emph{dominant} and \emph{non-dominant} hands (or arms) are used in~\cite{guiard_asymmetric_1987,kimmerle2010development} to describe their roles in asymmetrical bimanual tasks. 
Specifically, the non-dominant hand often stabilizes the object and sets a frame of reference defining the motion of the dominant hand.
In robotics, the leader-follower~\cite{Zhou2016a,liu_robot_2022} and \master{}-\slave{}~\cite{ureche_constraints_2018, gao_kvil_2023} relationships are also widely used to design control policies for the \slave{}/follower arm within a local frame defined on the \master{}/leader arm. 
In this paper, we adopt the \master{}-\slave{} relationship (MSR) naming convention following~\cite{ureche_constraints_2018, gao_kvil_2023},
and extend it from arm coordination to object relationships, while considering the human hands as a special type of object.
As a result, we unify the representation of the \emph{roles}, \emph{relationships}, and \emph{task constraints} for both objects and hands.
The bimanual manipulation categories~\cite{krebs_bimanual_2022} of the demonstrated tasks are then derived from the extracted MSR (see \cref{subsec:coordination}).
Overall, Bi-KVIL unifies the learning of object-centric uni- and bimanual manipulation tasks, and captures fine-grained manipulation styles.
To the best of our knowledge, this work is the first to simultaneously extract bimanual coordination strategies 
and generalizable geometric task constraints from few \mbox{($\sim$~5-10)} visual demonstrations.


The contributions of this paper are twofold: 
\begin{enumerate*}[label=(\roman*)]
    \item We propose Bi-KVIL
    for learning bimanual manipulation tasks from a small number of visual demonstrations. Bi-KVIL automatically extracts a \emph{Hybrid Master-Slave Relationship} (HMSR), the corresponding bimanual coordination strategy, and sub-symbolic task representations (see \cref{sec:bikvil}). These representations include keypoints-based geometric constraints on principal manifolds, their associated local frames, and movement primitives (see \cref{sec:background});
    \item We present the bimanual keypoint-based admittance controller (Bi-KAC) extended from KAC~\cite{gao_kvil_2023} to handle a set of prioritized geometric constraints for bimanual tasks (see \cref{sec:bikac}). 
    It allows the reproduction of bimanual tasks corresponding to the bimanual manipulation taxonomy introduced in~\cite{krebs_bimanual_2022}.
\end{enumerate*}


\section{Related Work}
\label{sec:relatedwork}

Learning fine-grained bimanual tasks from visual observation of human demonstrations is a long-standing goal in robotics. It combines challenges in computer vision, bimanual coordination, and control. Most previous works focus on one or a few aspects of the problem.


\subsection{Visual Imitation Learning}\label{sec:vil_related}

VIL has made impressive progress thanks to the advances in deep-learning-based computer vision algorithms. 
Perception pipelines~\cite{qin_dexmv_2022,patel_learning_2022} are used to obtain poses of hands and objects from visual demonstrations, which are then used to train reinforcement learning (RL) algorithms for motion policies. 
Despite their performance, generalization capabilities are not guaranteed in semantic manipulation~\cite{sundaresan_learning_2020} when objects have large shape variations.
To improve category-level generalization, visual object descriptors based on image features~\cite{florencemanuelli2018dense,deekshith_visual_2020,florence_self-supervised_2020,hadjivelichkov_oneshot_2022,amir_deep_2022,liu_gift_2019,yen2022nerfsupervision} were proposed to find dense correspondences between categorical objects, thus facilitating category-level adaptation of downstream object-centric manipulation skills.
Similarly, SE(3)-equivariant object shape features~\cite{simeonov_neural_2021,simeonov_se_2022,chun_local_2023a} and space coverage features~\cite{zhao_relationship_2016,huang_nift_2023} were proposed to cope with partially-observed object point-clouds.
However, VIL based on such features requires manually-annotated keypoints for training~\cite{sundaresan_kite_2023a,gao_kpam_2021} and inference~\cite{simeonov_se_2022}.
To address this issue, we adopted image features from~\cite{florencemanuelli2018dense} and proposed a \emph{Principal Constraint Estimation} (PCE) algorithm to automatically extract keypoints-based geometric constraints from demonstrations~\cite{gao_kvil_2023}. This approach outperforms data-driven methods~\cite{jin_geometric_2020_} in terms of the number of demonstrations and category-level generalization.  
However, considering bimanual coordination, it is crucial to apply any of these approaches to bimanual tasks.


\subsection{Imitation Learning of Bimanual Manipulation}

Many works on bimanual manipulation focus on designing controllers coping with known coordination categories~\cite{ajoudani_natural_2014,savic_hybrid_2016,almeida_lyapunovbased_2019,mirrazavisalehian_unified_2018,gao_projected_2018,park_extended_2015,lee_redundancy_2015,amadio_exploiting_2019} rather than learning the coordination strategies from demonstrations. Such strategies can be either implicitly encoded in the motions or explicitly represented as constraints.

\subsubsection{Implicit coordination} Trajectory-based bimanual imitation learning focuses on learning the spatio-temporal correlations of bilateral motions with different variations of movement primitives~\cite{pairet_learning_2019,franzese_interactive_2023,dong_passive_2022,knaust_guided_2021} or Transformer-based models~\cite{liu_robot_2022}. 
This implicit encoding of coordination strategies overlooks the roles of objects in the task, thus limiting generalization abilities compared to object-centric VIL approaches.
Coordination strategies are also implicitly encoded in bimanual deep imitation learning \cite{zhao_learning_2023,mobile_aloha,chen_humanlevel_2022,kataoka_bimanual_2022,xie_deep_2020,kim_robot_2022a,kim_transformerbased_2021}, which additionally requires many demonstrations that are not always available in the real world. 
Despite the success of their reactive controllers within the trained scenes,
these approaches lack generalization abilities 
as they do not explicitly encode coordination strategies and constraints. 
In contrast, Bi-KVIL only requires \mbox{5-10} demonstrations and improves generalization by explicitly extracting coordination strategies and task constraints.


\subsubsection{Explicit coordination}
Abstracting a representation of a coordinated behavior often involves analyzing the contact and grasp state, the role of the objects/hands, as well as the spatio-temporal and force constraints.
A rule-based classification was proposed in~\cite{krebs_bimanual_2022} to determine the category of bimanual actions defined by the bimanual manipulation taxonomy. 
Other works mainly focused on object-action relation~\cite{Dreher2020} or on learning a specific coordination strategy~\cite{ureche_constraints_2018}. 
In this paper, we focus on spatio-temporal constraints, as force data required by~\cite{ureche_constraints_2018} is not available in demonstration videos. 
Specifically, we unify the bimanual coordination categories of~\cite{krebs_bimanual_2022} in our MSR representation and controller.
Moreover, Bi-KVIL relaxes the need for predefined frames per object as in~\cite{ureche_constraints_2018}, and combines automatic extraction of MSR, bimanual coordination, and object-centric task representations, thus enabling generalizable fine-grained skills.



\section{Background}
\label{sec:background}
\begin{figure}[t]
    \centering
    \begin{subfigure}{0.31\columnwidth}
        \centering
        \begin{tikzpicture}
            \node (image) at (0,0) {
                    \includegraphics[height=30mm]{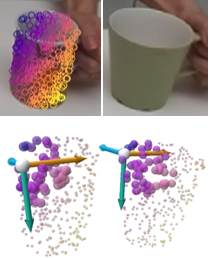}
                };
            \node (lf1) at (-1.15, -0.8) {\scalebox{0.7}{$\hat\lf$}};
            \node (lf2) at ( 0, -0.6) {\scalebox{0.7}{$\lf$}};
            \node (lf1) at (-0.7, -1.5) {\scalebox{0.6}{canonical}};
            \node (lf1) at ( 0.5, -1.5) {\scalebox{0.6}{demo${}_i$}};

            \coordinate (start) at (-0.5, -0.9);
            \coordinate (end) at (0.5, -0.9);
            \draw[-{Stealth[length=2mm]}] (start) to[out=-30,in=-150, line width=1.5pt, ] node[midway, below=0.1mm] {\scalebox{0.6}{DON}} (end);
        \end{tikzpicture}
        \caption{Local frame.}
        \label{subfig:kvil_review_frame}
    \end{subfigure}
    \begin{subfigure}{0.34\columnwidth}
        \centering\begin{tikzpicture}
            \node (image) at (0,0) {
                \includegraphics[height=30mm, trim=5 0 14 3, clip]{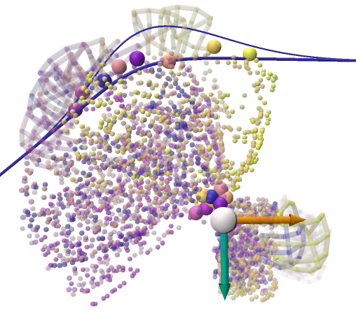}
                };
            \node (lf)  at ( 0.3, -0.8) {\scalebox{0.7}{$\lf$}};
            \node (p2p) at ( 0.8, -0.2) {\scalebox{0.7}{$\ptop$}};
            \node (p2c) at ( 1.2, 0.8) {\scalebox{0.7}{$\ptoc$}};
            \node (p2c) at ( 1.1, 1.2) {\scalebox{0.7}{density}};
            
            \node (lf)  at ( 0.8, -1.4) {\scalebox{0.7}{cup}};
            \node (p2c) at (-1.2, -1.4) {\scalebox{0.7}{kettle}};
            \node (lh)  at (-1.2,  1.2) {\scalebox{0.7}{\obj{rh}}};
            \node (rh)  at ( 1.2, -1.1) {\scalebox{0.7}{\obj{lh}}};
        \end{tikzpicture}
        \caption{Aligned view.}
        \label{subfig:kvil_review_align}
    \end{subfigure}
    \begin{subfigure}{0.31\columnwidth}
        \centering
        \begin{tikzpicture}
            \node (image) at (0,0) {
                \includegraphics[height=30mm]{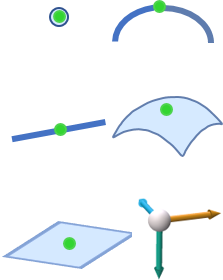}
            };
            \node (p2p) at (-0.55,  0.95) {\scalebox{0.7}{$\ptop$}};
            \node (p2l) at (-0.55, -0.25) {\scalebox{0.7}{$\ptol$}};
            \node (p2P) at (-0.55, -1.60) {\scalebox{0.7}{$\ptoP$}};
            \node (p2c) at ( 0.55,  0.95) {\scalebox{0.7}{$\ptoc$}};
            \node (p2S) at ( 0.55, -0.25) {\scalebox{0.7}{$\ptoS$}};
            \node (pose) at ( 0.55, -1.60) {\scalebox{0.7}{$\pose$}};
        \end{tikzpicture}
        \caption{Constraints.}
        \label{subfig:kvil_review_constr}
    \end{subfigure}
    \caption{Illustration of K-VIL on a cup-kettle \master{}-\slave{} pair. 
    }
    \label{fig:kvil_review}
    \vspace{-0.4cm}
\end{figure}

Here, we briefly review the K-VIL framework~\cite{gao_kvil_2023}, from which Bi-KVIL is derived.
Given a set of human demonstration videos (see \cref{subfig:teaser_demo}) with different categorical objects (\eg, different cups in \cref{subfig:kvil_review_frame}), 
K-VIL detects and tracks dense correspondence points, \ie, candidate points, on the visible surface of the objects using Dense Object Net (DON)~\cite{florencemanuelli2018dense}. The motion-salient object is considered a \slave{} object.
Candidate local frames are determined by matching the neighboring points between the canonical shape and actual point cloud of the \master{} object (see \cref{subfig:kvil_review_frame}).
We then align all demonstrations to each candidate local frame on the \master{} object. 
This allows the geometric constraints, \ie, spatial invariances, to become salient (see \cref{subfig:kvil_review_align}) and to be computed using the Principal Manifold Estimation (PME) algorithm~\cite{meng_principal_2021}.
In this way, the extracted local frame $\lf$, geometric constraints ($\ptop$ and $\ptoc$ in~\cref{subfig:kvil_review_align}), and associated via-point movement primitives (VMPs)~\cite{zhou_learning_2019} define the motion of the \slave{} object with respect to the \master{} object.
As shown in \cref{subfig:kvil_review_constr}, K-VIL considers, in priority order, point-to-point ($\ptop$), point-to-line ($\ptol$), point-to-plane ($\ptoP$), point-to-curve ($\ptoc$), and point-to-surface ($\ptoS$) geometric constraints. 
Each keypoint is then driven by a spring-damper system following the VMPs and constraints via a keypoint-based admittance controller (KAC). The composed force finally drives the robot hand to reproduce the task. 
Note that K-VIL focuses on a single \master{}-\slave{} pair, which corresponds to each \master{}-\slave{} pair in the Bi-KVIL HMSR graph (see \cref{subfig:teaser_msr} and \cref{fig:hmsr_diag}).

\section{Bimanual Keypoint-based Visual Imitation Learning (Bi-KVIL)}
\label{sec:bikvil}
This section presents the Bi-KVIL approach.
We first introduce the perception pipeline for preprocessing demonstration videos in \cref{subsec:perception}. We then detail the proposed \master{}-\slave{} relationship and its extraction in \cref{subsec:hms}.


\subsection{Preprocessing}
\label{subsec:perception}

(Bi-)K-VIL rely on robust estimation and tracking of dense candidate points of the objects. 
To this end, we compose various off-the-shell computer vision algorithms into a reliable perception pipeline. 
We support videos taken by stereo RGB or monocular RGB-D cameras from different viewpoints. 
We prefer the stereo approach for translucent or thin objects,
where the depth is estimated using UniMatch~\cite{xu_unifying_2023}.


\subsubsection{Candidate points on objects}

Similarly to K-VIL, we train DONs per object category in a self-supervised and task-agnostic manner. 
To improve the data quality, we replace the traditional 3D reconstruction in DON with a modified Instant-NGP following~\cite{muller_instant_2022,ichnowski_dexnerf_2021}.
We select the first image frame of a random demonstration to create the canonical space of all objects (see \cref{subfig:kvil_review_frame}). 
We use DON to find the initial dense correspondence points for each object \emph{only} on the first image frame of any other demonstrations, and the deep optical flow algorithm RAFT~\cite{teed_raft_2020} to track the motion of these points in image coordinates. 
To further restrict the results of DON and RAFT within the region of the objects, we employ the Segment~\cite{kirillov_segment_2023} and Track~\cite{cheng_segment_2023} Anything models in combination to the object detection model Grounding DINO~\cite{liu_grounding_2023}.
Finally, we map the motion of the candidate points from image coordinates to 3D using triangular geometry, remove outliers, and smooth the motions with Savitzky-Golay filters.


\subsubsection{Human pose estimation} 

In addition to candidate points on objects, Bi-KVIL requires keypoints of the human hands and the handedness, \ie, the left/right label of each hand. 
In natural visual demonstrations, human hands are often heavily (self-)occluded in several image frames, where methods like MediaPipe~\cite{lugaresi_mediapipe_2019} fail. 
We found that RTMPose~\cite{jiang_rtmpose_2023} robustly estimates the whole-body human pose in 2D including the handedness, which allows us to map different sub-tasks to the robot's hands. 
The image patches containing the detected hands and their handedness are used to obtain 3D hand poses using MeshGraphormer~\cite{lin_mesh_2021}. 
We re-base the hand mesh to the most probable visible keypoint of the hand using object-hand mask overlay and pre-defined priority. 
We empirically observed that our framework outperforms other models, \eg, MediaPipe 3D, OSX~\cite{lin_onestage_2023a} when under heavy (self-)occlusion.

It is important to note that we assume the human demonstrations to be temporally segmented, so we focus on extracting the HMSR and K-VIL's task representation for each motion segment. 
We do not include evaluations of different computer algorithms, since this is not the focus of our paper.


\begin{figure*}[t]
    \centering
    \begin{subfigure}{0.1\textwidth}
        \centering
        \begin{tikzpicture}
          \node[draw, fill=blue!15, ellipse] (Base) at (0, 0) {\mast{0}{0}};
        
          \node[draw, fill=red!15, ellipse] (Derived1) at (0, -1.7) {\slav{0}};
        
          \draw[-latex] (Derived1) -- (Base) node[midway, left] {};
        \end{tikzpicture}
        \caption{Single.}
        \label{subfig:hmsr_single}
    \end{subfigure}
    \begin{subfigure}{0.19\textwidth}
        \centering
        \begin{tikzpicture}
          \node[draw, fill=blue!15, ellipse] (Base1) at (-0.75, 0) {\mast{0}{0}};
          \node[draw, fill=blue!15, ellipse] (Base2) at ( 0.75, 0) {\mast{0}{1}};
        
          \node[draw, fill=red!15, ellipse] (Derived1) at (0, -1.7) {\slav{0}};
        
          \draw[-latex] (Derived1) -- (Base1) node[midway, left] {};
          \draw[-latex] (Derived1) -- (Base2) node[midway, right] {};
        \end{tikzpicture}
        \caption{Multiple.}
        \label{subfig:hmsr_multiple}
    \end{subfigure}
    \begin{subfigure}{0.12\textwidth}
        \centering
        \begin{tikzpicture}
          \node[draw, fill=blue!15, ellipse] (Master) at (0, 0) {\mast{0}{0}};
          \node[draw, fill=green!15, ellipse] (Middle) at (0, -0.85) {\mast{1}{0}};
          \node[draw, fill=red!15, ellipse] (Slave) at (0, -1.7) {\slav{0}};          
        
          \draw[-latex] (Slave) -- (Middle) node[midway, left] {};
          \draw[-latex] (Middle) -- (Master) node[midway, right] {};
        \end{tikzpicture}
        \caption{Multi-level.}
        \label{subfig:multi_level}
    \end{subfigure}
    \begin{subfigure}{0.25\textwidth}
        \centering
        \begin{tikzpicture}
          \node[draw, fill=blue!15, ellipse] (m0) at (0, 0) {\mast{0}{0}};       
          \node[draw, fill=green!15, ellipse] (m1_0) at (-1.0, -0.85) {\mast{1}{0}};       
          \node[draw, fill=green!15, ellipse] (m1_1) at ( 1.0, -0.85) {\mast{1}{1}};
          \node[draw, fill=red!15, ellipse] (s0) at (-1.50, -1.7) {\slav{0}}; 
          \node[draw, fill=red!15, ellipse] (s1) at (-0.50, -1.7) {\slav{1}}; 
          \node[draw, fill=red!15, ellipse] (s2) at ( 0.50, -1.7) {\slav{2}}; 
          \node[draw, fill=red!15, ellipse] (s3) at ( 1.50, -1.7) {\slav{3}}; 
        
          \draw[-latex] (s0) -- (m1_0) node[midway, left] {};
          \draw[-latex] (s1) -- (m1_0) node[midway, left] {};
          \draw[-latex] (s2) -- (m1_1) node[midway, left] {};
          \draw[-latex] (s3) -- (m1_1) node[midway, left] {};
          
          \draw[-latex] (m1_0) -- (m0) node[midway, left] {};
          \draw[-latex] (m1_1) -- (m0) node[midway, left] {};
        \end{tikzpicture}
        \caption{Hierarchical.}
        \label{subfig:hmsr_hier}
    \end{subfigure}
    \begin{subfigure}{0.25\textwidth}
        \centering
        \begin{tikzpicture}
          \node[draw, fill=blue!15, ellipse] (m0_0) at (-1.75, 0) {\mast{0}{0}};  
          \node[draw, fill=blue!15, ellipse] (m0_1) at (0, 0) {\mast{0}{1}};   
          \node[draw, fill=blue!15, ellipse] (m0_2) at (1.5, 0) {\mast{0}{2}}; 
          
          \node[draw, fill=green!15, ellipse] (m1_0) at (-1.0, -0.85) {\mast{1}{0}};       
          \node[draw, fill=green!15, ellipse] (m1_1) at ( 1.0, -0.85) {\mast{1}{1}};
          \node[draw, fill=red!15, ellipse] (s0) at (-1.50, -1.7) {\slav{0}}; 
          \node[draw, fill=red!15, ellipse] (s1) at (-0.50, -1.7) {\slav{1}}; 
          \node[draw, fill=red!15, ellipse] (s2) at ( 0.50, -1.7) {\slav{2}}; 
          \node[draw, fill=red!15, ellipse] (s3) at ( 1.50, -1.7) {\slav{3}}; 
        
          \draw[-latex] (s0) -- (m1_0) node[midway, left] {};
          \draw[-latex] (s1) -- (m1_0) node[midway, left] {};
          \draw[-latex] (s2) -- (m1_1) node[midway, left] {};
          \draw[-latex] (s3) -- (m1_1) node[midway, left] {};
          
          \draw[-latex] (s0) -- (m0_0) node[midway, left] {};
          \draw[-latex, blue, dashed, line width=0.8pt, dash pattern=on 2pt off 1pt] (s1) -- (m0_1) node[midway, left] {};
          \draw[-latex] (m1_0) -- (m0_1) node[midway, left] {};
          \draw[-latex] (m1_1) -- (m0_2) node[midway, left] {};
          \draw[-latex] (m1_1) -- (m0_1) node[midway, left] {};
        \end{tikzpicture}
        \caption{Hybrid.}
        \label{subfig:hmsr_hybrid}
    \end{subfigure}
    \caption{MSR diagrams. $\mathsf{M}^{i}_{j}$ represents the $j$-th \master{} object at level $i$. For level $i>0$, a \master{} object is itself a \slave{} object paired to another \master{} object at level $i-1$. The \slave{} objects are located at the lowest level.}
    \label{fig:hmsr_diag}
    \vspace{-0.5cm}
\end{figure*}
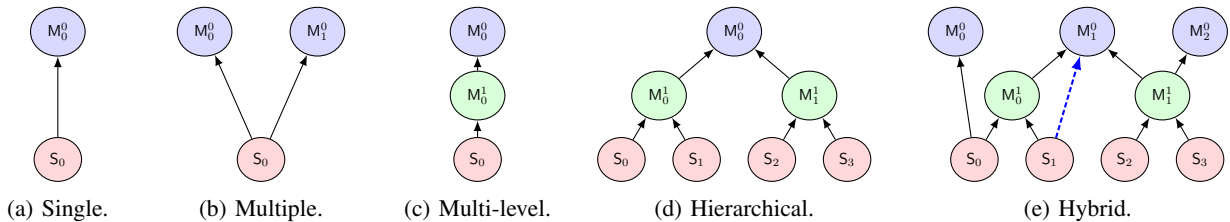


\subsection{Extraction of Master-Slave Relationships (MSRs)}
\label{subsec:hms}

Given the 3D trajectories of dense points on the objects and hands aligned in the camera frame, we first analyze the object/hands relationships.
We propose five types of MSRs that often appear in unimanual and bimanual manipulation tasks, namely, \emph{single, multiple, multi-level, hierarchical}, and \emph{hybrid} MSR, which resemble the definition and graph representation of the inheritance in C++ programming language (see \cref{fig:hmsr_diag}).
The single MSR corresponds to the unimanual K-VIL case~\cite{gao_kvil_2023}, where only two objects interact within a single \master{}-\slave{} pair. 
Within multiple MSR, the motion of a \slave{} object is defined in local frames of multiple \master{} objects. 
In multi-level MSR, a \slave{} object can be the \master{} of another \slave{} object. 
A hierarchical MSR is a tree-like structure where each \master{} may have multiple \slave{} objects.
Finally, the hybrid MSR (HMSR) combines multiple and hierarchical MSRs, \ie,
it is a directed acyclic graph (DAG) in which a slave object may have master objects at different levels (\eg, \scalebox{1.3}{$\slav{1}$} in \cref{subfig:hmsr_hybrid}).
The HMSR differs from the inheritance in C++ in that a \slave{} object does not inherit geometric constraints from its \master{}. 
Instead, constraints are explicitly defined between each object pair (see \inlinearrow{dashed, blue, line width=1.5pt, dash pattern=on 2pt off 1pt} in \cref{subfig:hmsr_hybrid}). 
In the following, we use HMSR as a general framework that encompasses all other MSRs.
To extract HMSR, we first build a rough DAG using motion-saliency, grasping, and pose invariance detection in \cref{subsec:abs_saliency,subsec:rel_saliency,subsec:pose_invariance}.
Since each valid \master{}-\slave{} pair in this graph must have at least one of the constraints in \cref{subfig:kvil_review_constr}, we use \mbox{K-VIL} in \cref{subsec:truncate} to truncate \master{}-\slave{} pairs without constraints and finally obtain a compact graph.


\subsubsection{Absolute motion saliency detection}
\label{subsec:abs_saliency}
Since any motion must be represented in a local frame, the top-level master objects in HMSR must be static.
We use motion saliency to determine \emph{static} (if any) and \emph{moving} objects based on the average point velocities in the global camera frame.
In the absence of static objects, we define local frames on the \emph{initial state} of the moving objects, 
in which the constraints and its motion following that state 
 are modeled (see \cref{subsec:truncate}).
We define the initial state, \ie, the observed point cloud at the first timestep, of each moving object as a special object, called \emph{virtual object}, in HMSR (see, \eg,~\cref{subfig:teaser_msr}).


\subsubsection{Grasp detection based on relative motion saliency}
\label{subsec:rel_saliency}
Before determining bimanual coordination strategies, we need to estimate the grasping relationships between hands and objects.
Similarly to K-VIL, the human hand is considered a special type of object with 21 keypoints (see MANO~\cite{romero_embodied_2017} model). 
We detect contacts between two objects based on the spatial distances between all candidate point pairs. 
Additionally, we compute the average change rate of the absolute distance of the $Q=50$ neighboring points on the object around the hand relative to the hand's local frame.
If it drops below a certain threshold for a hand-object pair in contact, a \emph{firm grasp} is detected.
In object-centric representations, the \emph{grasps} are modeled and adapted with respect to the object being grasped. 
Therefore, we set the hand as a \slave{} object paired to the grasped \master{} object (see~\cref{subfig:teaser_msr}).


\subsubsection{Pose Invariance Detection}
\label{subsec:pose_invariance}

Our approach relies on the estimated motion of the objects' candidate points without any prior semantic knowledge about the objects or their roles in the task. 
Therefore, the motion of object $\mathsf{A}$ relative to $\mathsf{B}$ can equivalently be represented as the motion of $\mathsf{B}$ relative to $\mathsf{A}$.
This results in a potential bi-directional MSR,  leading to improper reproductions. For example, the \master{} cup may move with respect to the \slave{} kettle, resulting in an invalid pouring action. 
To address this issue, Krebs et. al.~\cite{krebs_bimanual_2022} chose the \master{} object as the less mobile one using absolute motion saliency detection. 
However, this is not necessarily correct. 
For example, the cup is usually considered a \master{} object in the pouring task even if it
moves more than the \slave{} kettle (see \cref{subfig:teaser_constr}).
To address this problem, we propose an \emph{invariance criterion}.
Given a moving object pair $(O_\mathsf{A}, O_\mathsf{B})$ that has a potential bi-directional MSR, we first compute the translational and orientational spatial invariance of both relative to all static objects $\set{O_s}$.
The key idea is that if we observe the most salient spatial invariance from object $O_l$ with respect to $O_s$, $l\in\set{\mathsf{A}, \mathsf{B}}$, we consider $O_l$ the \master{}, which itself is paired to the \master{} $O_s$.
Specifically, the \master{} object is obtained by
\begin{equation}
    \textstyle l = \argmin_l \set{r^p_{s,l}, r^o_{s,l}}_{s \in \mathcal{O}_{s}, \;l \in \set{\mathsf{A}, \mathsf{B}}},
\end{equation}
where the ratios $r^p_{s,l}, r^o_{s,l}$ are the normalized translational and orientational spatial variability of the salient object $O_l$ relative to the static object $O_s$, respectively. 
This ensures that the HMSR is a DAG.
The HMSR may still contain redundant relations, which we then truncate. 


\subsubsection{Truncation}\label{subsec:truncate}

For each potential \master{}-\slave{} pair in the HMSR graph, we employ K-VIL as described in \cref{sec:background} and remove the pairs that do not show any constraint between the \master{} and \slave{} objects. 
This results in a compact HMSR graph associated with sub-symbolic geometric constraints for each \master{}-\slave{} pair (see~\cref{sec:background} and~\cref{subfig:teaser_msr,subfig:teaser_constr}).
In \cref{sec:eval}, we show that the HMSR graph becomes more compact and converges as the number of demonstrations increases. 
Our insight is that, with scarce demonstrations, any valid salient geometric constraint should be considered as knowledge about the task is limited, while unnecessary constraints can be truncated when statistical evidence becomes available in new demonstrations.
When a moving \master{} object $O_{\mathsf{m}}$ is not constrained by any static object after truncation, it is allowed to move freely in space following a task-space VMP. 
Its pose corresponds to the pose of the local frame that defines the top-priority constraints for its \slave{} object.
The frame of reference for the VMP is located on the virtual object, defined by the initial state of object $O_{\mathsf{m}}$.
We model a distribution of the end pose of $O_{\mathsf{m}}$ in the demonstrations, from which we sample a target pose to adapt the VMP for execution.
This new type of constraint, called pose constraint, is added to  \cref{subfig:kvil_review_constr}.


\begin{table*}
    \centering
    \setlength\tabcolsep{0.5pt}
    \setlength\imgwidth{19mm}
    \begin{tabular}{cM{\imgwidth}M{\imgwidth}M{22mm}M{22mm}M{\imgwidth}M{\imgwidth}M{\imgwidth}M{\imgwidth}}
       \toprule
            Tasks 
            & \task{Pl^{1}_{sp}}{3}
            & \task{Pl^{1}_{sp}}{4} 
            & \task{Pl^{1}_{sp}}{5} 
            & \task{Pl^{1}_{sp}}{6} 
            & \task{Pl^{2}_{sp}}{6}
            & \task{Pl^{3}_{sp}}{6}
            & \task{Pl^{4}_{sp}}{6}
            & \task{Pl^{5}_{sp}}{6}
        \\
        \midrule
            \multirow{1}{*}[2em]{\rotatebox[origin=c]{90}{Demo.}}
            & \multicolumn{4}{c|}{\includegraphics[height=10mm, trim=200 0 120 420, clip]{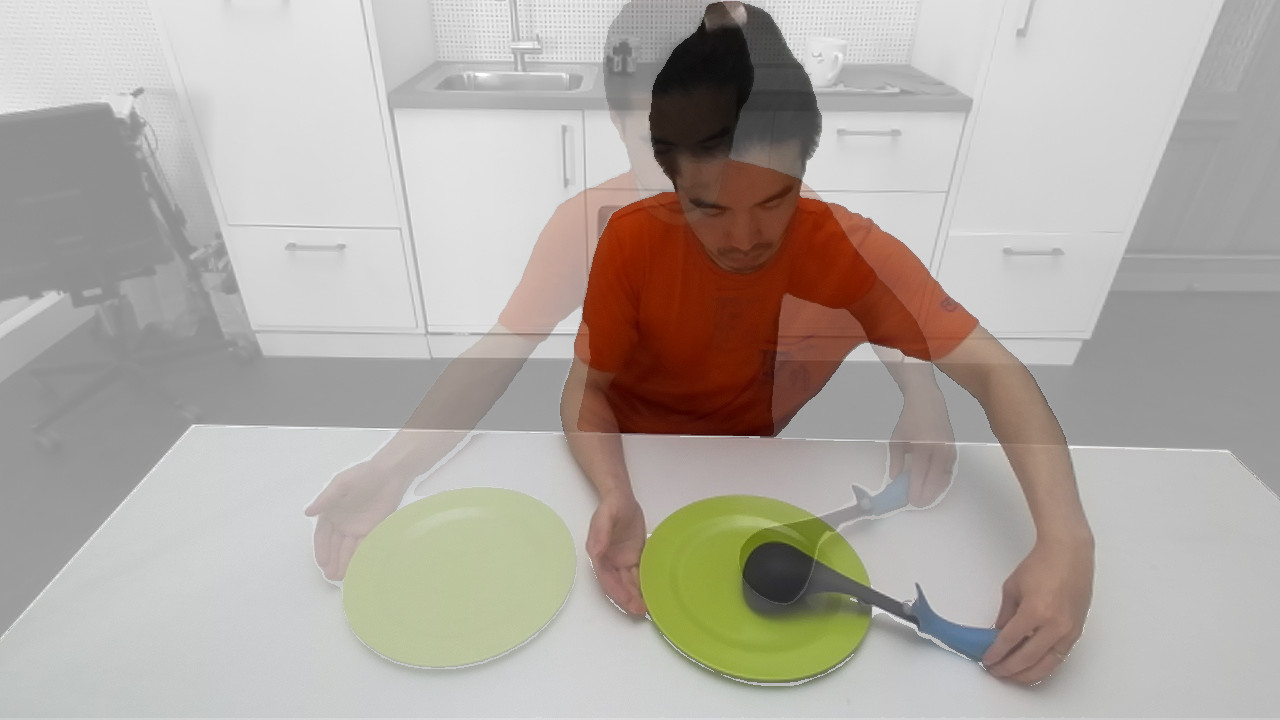}}
            & \multicolumn{1}{c}{\includegraphics[height=10mm, trim=280 0 220 220, clip]{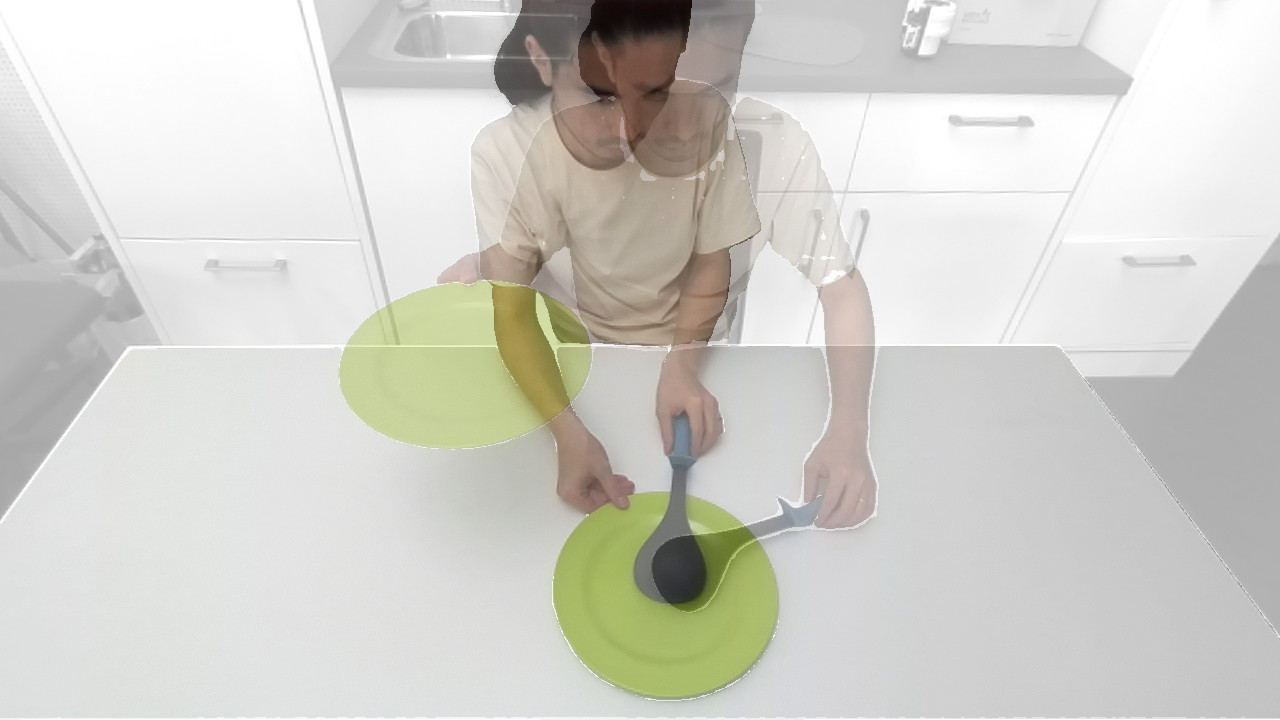}}
            & \multicolumn{1}{c}{\includegraphics[height=10mm, trim=240 90 350 220, clip]{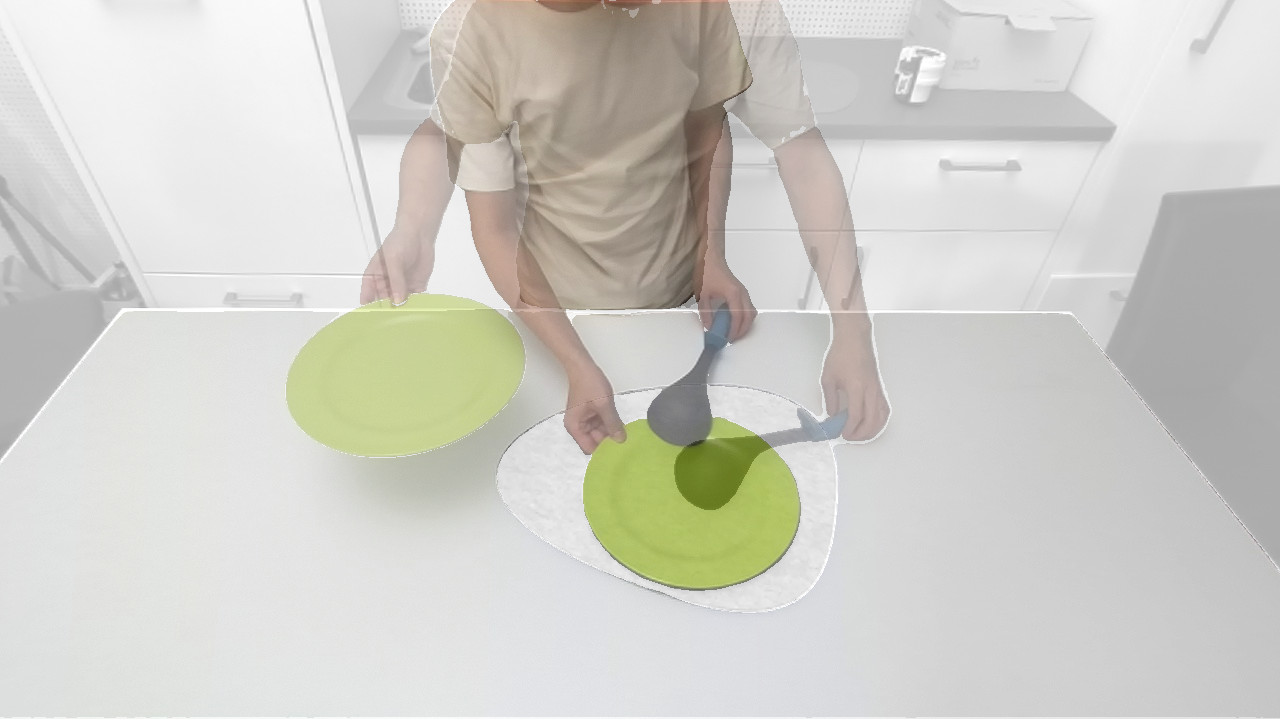}}
            & \multicolumn{1}{c}{\includegraphics[height=10mm, trim=400 0 200 400, clip]{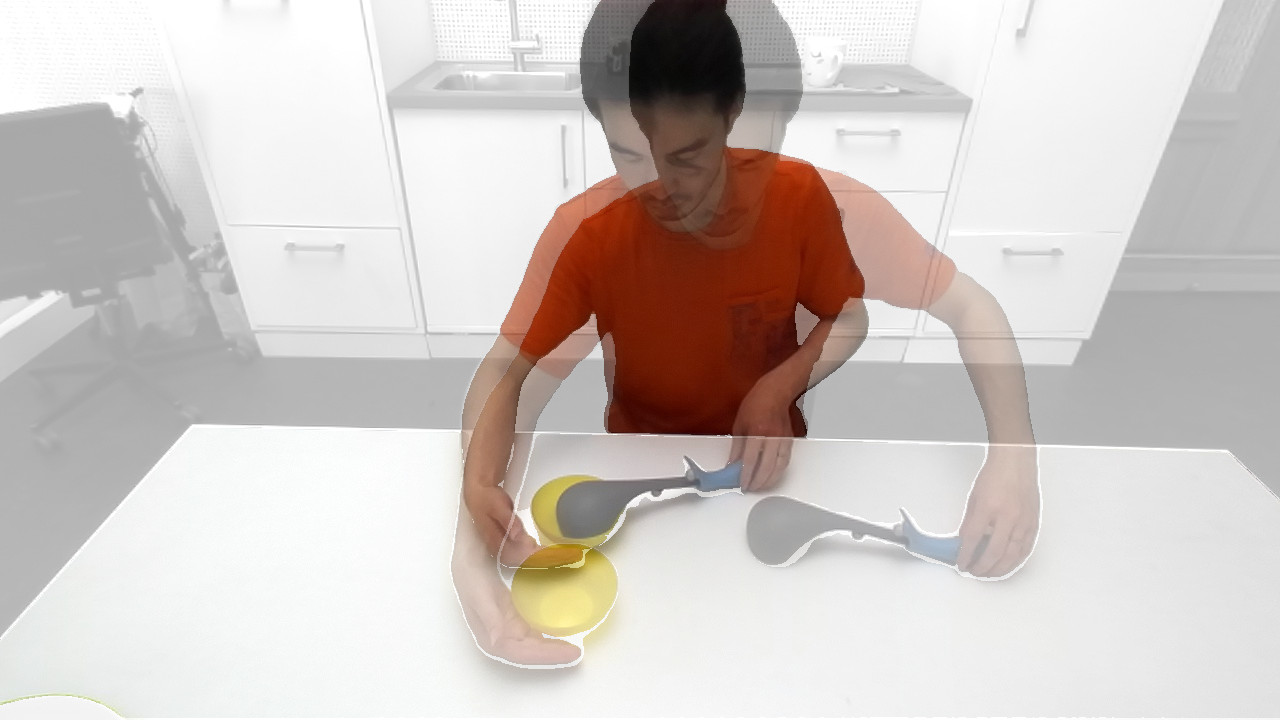}}
            & \multicolumn{1}{c}{\includegraphics[height=10mm, trim=280 90 250 250, clip]{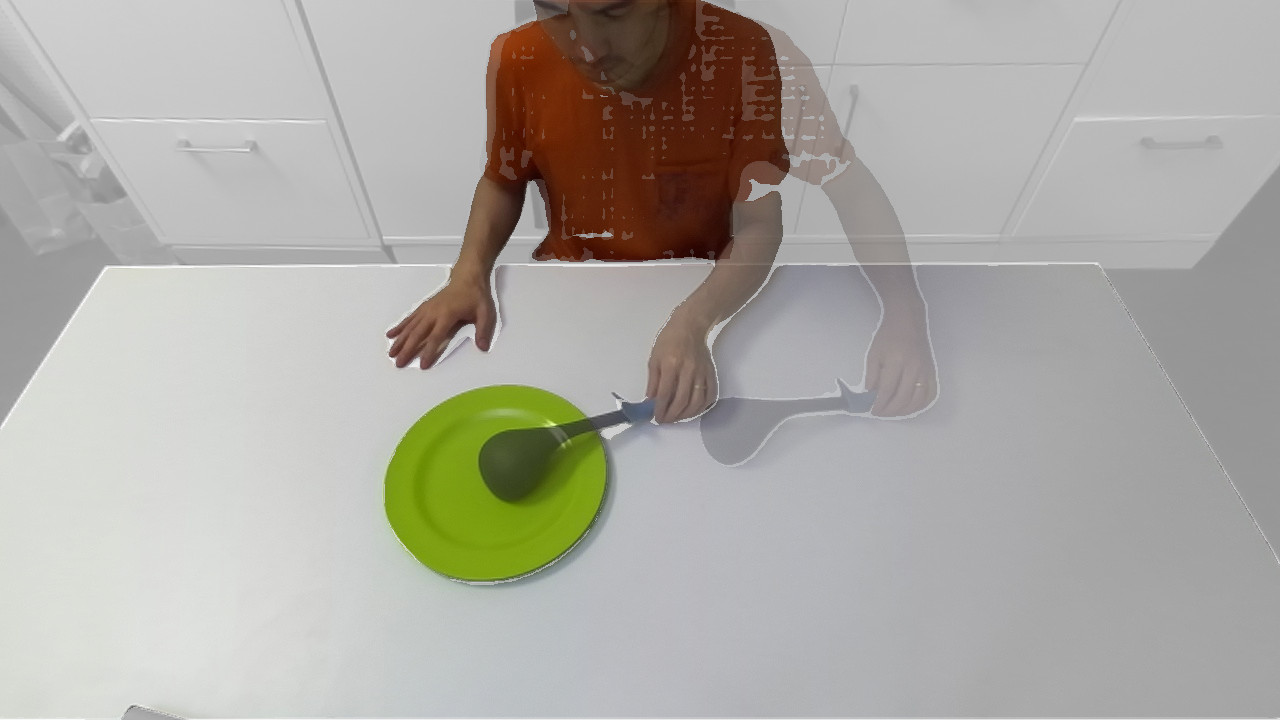}}
        \\
            \multirow{1}{*}[1em]{\rotatebox[origin=c]{90}{HMSR}}
            & \resizebox{!}{27mm}{\begin{tikzpicture}
  \node[rectangle, rounded corners, minimum width=9mm, minimum height=4mm, fill=gray!20] at (-0.90, -0.34) {};
  
  \node[draw, fill=blue!15,  ellipse, minimum width=8mm, minimum height=7mm] (vs) at ( 0.0,  0.0) {\obj{vs}};       
  \node[draw, fill=green!15, ellipse, minimum width=8mm, minimum height=7mm] (p)  at (-1.0, -1.0) {\obj{pt}};       
  \node[draw, fill=green!15, ellipse, minimum width=8mm, minimum height=7mm] (s)  at (-0.0, -2.0) {\obj{sp}};
  
  \node[draw, fill=red!15,   ellipse, minimum width=8mm, minimum height=7mm] (lh) at ( 0.0, -3.0) {\obj{lh}};
  \node[draw, fill=red!15,   ellipse, minimum width=8mm, minimum height=7mm] (rh) at (-1.0, -3.0) {\obj{rh}};

  \draw[-latex, blue] (p) -- (vs) node[pos=0.7, left]         {\scalebox{0.9}{2\obj{\times p2p}}};
  \draw[-latex, blue] (s) -- (p)  node[pos=0.1, left=0.5mm]   {\scalebox{0.9}{\obj{p2p}}};
  \draw[-latex, blue] (s) -- (vs) node[pos=0.5, right=0.1mm]  {\scalebox{0.9}{\obj{p2p}}};
  \draw[-latex, orange] (lh) -- (s)  node[pos=0.5, right=0.1mm]  {\scalebox{0.9}{\obj{g}}};
  \draw[-latex, orange] (rh) -- (p)  node[pos=0.13, right=0.1mm] {\scalebox{0.9}{\obj{g}}};
  \draw[-latex, orange] (lh) -- (s)  node[pos=0.5, right=0.1mm]  {\scalebox{0.9}{\obj{g}}};
  \draw[-latex, orange] (rh) -- (p)  node[pos=0.13, right=0.1mm] {\scalebox{0.9}{\obj{g}}};
  
\end{tikzpicture}}
            & \resizebox{!}{27mm}{\begin{tikzpicture}
  \node[rectangle, rounded corners, minimum width=9mm, minimum height=4mm, fill=gray!20] at (-0.90, -0.34) {};
  \node[draw, fill=blue!15,  ellipse, minimum width=8mm, minimum height=7mm] (vs) at ( 0.0,  0.0) {\obj{vs}};       
  \node[draw, fill=green!15, ellipse, minimum width=8mm, minimum height=7mm] (p)  at (-1.0, -1.0) {\obj{pt}};       
  \node[draw, fill=green!15, ellipse, minimum width=8mm, minimum height=7mm] (s)  at (-0.0, -2.0) {\obj{sp}};
  
  \node[draw, fill=red!15,   ellipse, minimum width=8mm, minimum height=7mm] (lh) at ( 0.0, -3.0) {\obj{lh}};
  \node[draw, fill=red!15,   ellipse, minimum width=8mm, minimum height=7mm] (rh) at (-1.0, -3.0) {\obj{rh}};

  \draw[-latex, blue] (p) -- (vs) node[pos=0.7, left]       {\scalebox{0.9}{2\obj{\times p2p}}};
  \draw[-latex, blue] (s) -- (p)  node[pos=0.1, left=0.1mm]   {\scalebox{0.9}{\obj{p2p}}};
  \draw[-latex, blue] (s) -- (vs) node[pos=0.7, right=0.1mm]  {\scalebox{0.9}{\obj{\ptop}}};
  \draw[-latex, blue] (s) -- (vs) node[pos=0.4, right=0.1mm]  {\scalebox{0.9}{2\obj{\times}}};
  \draw[-latex, blue] (s) -- (vs) node[pos=0.2, right=0.1mm]  {\scalebox{0.9}{\obj{\ptol}}};
  \draw[-latex, orange] (lh) -- (s)  node[pos=0.5, right=0.1mm]  {\scalebox{0.9}{\obj{g}}};
  \draw[-latex, orange] (rh) -- (p)  node[pos=0.13, right=0.1mm]  {\scalebox{0.9}{\obj{g}}};
  \draw[-latex, orange] (lh) -- (s)  node[pos=0.5, right=0.1mm]  {\scalebox{0.9}{\obj{g}}};
  \draw[-latex, orange] (rh) -- (p)  node[pos=0.13, right=0.1mm]  {\scalebox{0.9}{\obj{g}}};
  
\end{tikzpicture}}
            & \resizebox{!}{27mm}{\begin{tikzpicture}
  \node[draw=red, fill=blue!15, ellipse, minimum width=8mm, minimum height=7mm, dotted, dash pattern=on 2pt off 1pt, line width=0.8pt] (vp)  at (-1.2,  0.0) {\obj{vp}};  
  \node[draw, fill=blue!15, ellipse, minimum width=8mm, minimum height=7mm] (vs)  at ( 0.0,  0.0) {\obj{vs}}; 
  
  \node[draw, fill=green!15, ellipse, minimum width=8mm, minimum height=7mm] (p)  at (-1.2, -1.0) {\obj{pt}};       
  \node[draw, fill=green!15, ellipse, minimum width=8mm, minimum height=7mm] (s)  at (-0.0, -2.0) {\obj{sp}};
  
  \node[draw, fill=red!15,   ellipse, minimum width=8mm, minimum height=7mm] (lh) at ( 0.0, -3.0) {\obj{lh}};
  \node[draw, fill=red!15,   ellipse, minimum width=8mm, minimum height=7mm] (rh) at (-1.2, -3.0) {\obj{rh}};

  \draw[-latex, blue] (p) -- (vs) node[pos=1.1, left=0.5mm]   {\scalebox{0.9}{\obj{p2p}}};
  \draw[-latex, blue] (s) -- (vs) node[pos=0.6, right=0.1mm]  {\scalebox{0.9}{\obj{p2p}}};
  \draw[-latex, blue] (s) -- (vs) node[pos=0.4, right=0.1mm]  {\scalebox{0.9}{\obj{p2l}}};
  
  \draw[-latex, red, dashed, dash pattern=on 2pt off 1pt] (s) -- (vp) node[pos=0.7, right=1.2mm]  {\scalebox{0.9}{2\obj{\times}}};
  \draw[-latex, red, dashed, dash pattern=on 2pt off 1pt] (s) -- (vp) node[pos=0.5, right=-0.3mm]  {\scalebox{0.9}{\obj{p2P}}};
  \draw[-latex, red, dashed, dash pattern=on 2pt off 1pt] (p) -- (vp) node[pos=0.5, left=0.1mm]  {\scalebox{0.9}{\obj{p2P}}};
  
  \draw[-latex, blue] (s) -- (p)  node[pos=0.3, left=0.5mm]   {\scalebox{0.9}{\obj{p2p}}};
  \draw[-latex, blue] (s) -- (p)  node[pos=-0.1, left=0.8mm]   {\scalebox{0.9}{2\obj{\times p2P}}};
  
  \draw[-latex, orange] (lh) -- (s)  node[pos=0.5, right=0.1mm]  {\scalebox{0.9}{\obj{g}}};
  \draw[-latex, orange] (rh) -- (p)  node[pos=0.13, right=0.1mm] {\scalebox{0.9}{\obj{g}}};
  \draw[-latex, orange] (lh) -- (s)  node[pos=0.5, right=0.1mm]  {\scalebox{0.9}{\obj{g}}};
  \draw[-latex, orange] (rh) -- (p)  node[pos=0.13, right=0.1mm] {\scalebox{0.9}{\obj{g}}};
  
\end{tikzpicture}}
            & \resizebox{!}{27mm}{\begin{tikzpicture}
  \node[draw=red, fill=blue!15, ellipse, minimum width=8mm, minimum height=7mm, dashed, dash pattern=on 2pt off 1pt, line width=0.8pt] (vp)  at (-1.2,  0.0) {\obj{vp}};  
  \node[draw, fill=blue!15, ellipse, minimum width=8mm, minimum height=7mm] (vs)  at ( 0.0,  0.0) {\obj{vs}}; 
  
  \node[draw, fill=green!15, ellipse, minimum width=8mm, minimum height=7mm] (p)  at (-1.2, -1.0) {\obj{pt}};       
  \node[draw, fill=green!15, ellipse, minimum width=8mm, minimum height=7mm] (s)  at (-0.0, -2.0) {\obj{sp}};
  
  \node[draw, fill=red!15,   ellipse, minimum width=8mm, minimum height=7mm] (lh) at ( 0.0, -3.0) {\obj{lh}};
  \node[draw, fill=red!15,   ellipse, minimum width=8mm, minimum height=7mm] (rh) at (-1.2, -3.0) {\obj{rh}};

  \draw[-latex, blue] (p) -- (vs) node[pos=1.1, left=0.5mm]   {\scalebox{0.9}{\obj{p2p}}};
  \draw[-latex, blue, red, dashed, dash pattern=on 2pt off 1pt] (s) -- (vp) node[pos=0.7, right=1.2mm]  {\scalebox{0.9}{2\obj{\times}}};
  \draw[-latex, blue, red, dashed, dash pattern=on 2pt off 1pt] (s) -- (vp) node[pos=0.5, right=-0.3mm]  {\scalebox{0.9}{\obj{p2P}}};
  
  \draw[-latex, blue] (s) -- (p)  node[pos=0.1, left=0.5mm]   {\scalebox{0.9}{\obj{p2p}}};
  \draw[-latex, blue] (s) -- (vs) node[pos=0.6, right=0.1mm]  {\scalebox{0.9}{\obj{p2p}}};
  \draw[-latex, blue] (s) -- (vs) node[pos=0.4, right=0.1mm]  {\scalebox{0.9}{\obj{p2P}}};
  \draw[-latex, orange] (lh) -- (s)  node[pos=0.5, right=0.1mm]  {\scalebox{0.9}{\obj{g}}};
  \draw[-latex, orange] (rh) -- (p)  node[pos=0.13, right=0.1mm] {\scalebox{0.9}{\obj{g}}};
  \draw[-latex, orange] (lh) -- (s)  node[pos=0.5, right=0.1mm]  {\scalebox{0.9}{\obj{g}}};
  \draw[-latex, orange] (rh) -- (p)  node[pos=0.13, right=0.1mm] {\scalebox{0.9}{\obj{g}}};
  
\end{tikzpicture}}
            & \resizebox{!}{27mm}{\begin{tikzpicture}
  \node[draw, fill=blue!15, ellipse, minimum width=8mm, minimum height=7mm] (vs) at ( 0.0,  0.0) {\obj{vs}};       
  \node[draw, fill=green!15, ellipse, minimum width=8mm, minimum height=7mm] (p) at (-1.0, -1.0) {\obj{pt}};       
  \node[draw, fill=green!15, ellipse, minimum width=8mm, minimum height=7mm] (s) at (-0.0, -2.0) {\obj{sp}};
  
  \node[draw, fill=red!15, ellipse, minimum width=8mm, minimum height=7mm] (lh) at ( 0.0, -3) {\obj{lh}};
  \node[draw, fill=red!15, ellipse, minimum width=8mm, minimum height=7mm] (rh) at (-1.0, -3) {\obj{rh}};

  \draw[-latex, blue] (p) -- (vs) node[pos=1.0,  left=0.5mm]   {\scalebox{0.9}{\obj{p2p}}};
  \draw[-latex, blue] (s) -- (vs) node[pos=0.7,  right=0.1mm]  {\scalebox{0.9}{\obj{\ptop}}};
  \draw[-latex, blue] (s) -- (vs) node[pos=0.4,  right=0.1mm]  {\scalebox{0.9}{2\obj{\times}}};
  \draw[-latex, blue] (s) -- (vs) node[pos=0.2,  right=0.1mm]  {\scalebox{0.9}{\obj{\ptoP}}};
  \draw[-latex, blue] (s) -- (p)  node[pos=0.3,  left=0.5mm]   {\scalebox{0.9}{\obj{p2p}}};
  \draw[-latex, blue] (s) -- (p)  node[pos=-0.2, left=1mm]   {\scalebox{0.9}{\obj{p2P}}};
  \draw[-latex, orange] (lh) -- (s)  node[pos= 0.5, right=0.1mm]  {\scalebox{0.9}{\obj{g}}};
  \draw[-latex, orange] (rh) -- (p)  node[pos=0.13, right=0.1mm]  {\scalebox{0.9}{\obj{g}}};
  
  \draw[-latex, orange] (lh) -- (s)  node[pos= 0.5, right=0.1mm]  {\scalebox{0.9}{\obj{g}}};
  \draw[-latex, orange] (rh) -- (p)  node[pos=0.13, right=0.1mm]  {\scalebox{0.9}{\obj{g}}};
  
\end{tikzpicture}}
            & \resizebox{!}{27mm}{\begin{tikzpicture}
  \node[rectangle, rounded corners, minimum width=6mm, minimum height=6mm, fill=amber!30] at (-0.68, -1.72) {};
  \node[rectangle, rounded corners, minimum width=6mm, minimum height=6mm, fill=amber!30] at ( 0.30, -1.00) {};
  
  \node[draw, fill=blue!15, ellipse, minimum width=8mm, minimum height=7mm] (vs) at ( 0.0,  0.0) {\obj{vs}};       
  \node[draw, fill=green!15, ellipse, minimum width=8mm, minimum height=7mm] (p) at (-1.0, -1.0) {\obj{pt}};       
  \node[draw, fill=green!15, ellipse, minimum width=8mm, minimum height=7mm] (s) at (-0.0, -2.0) {\obj{sp}};
  
  \node[draw, fill=red!15, ellipse, minimum width=8mm, minimum height=7mm] (lh) at ( 0.0, -3) {\obj{lh}};
  \node[draw, fill=red!15, ellipse, minimum width=8mm, minimum height=7mm] (rh) at (-1.0, -3) {\obj{rh}};

  \draw[-latex, blue] (p) -- (vs) node[pos=0.7, left]      {\scalebox{0.9}{\obj{p2p}}};
  \draw[-latex, blue] (s) -- (vs) node[pos=0.5, right=0.1mm] {\scalebox{0.9}{\obj{\ptoP}}};
  \draw[-latex, blue] (s) -- (p)  node[pos=0.3, left=0.2mm]  {\scalebox{0.9}{2\obj{\times}}};
  \draw[-latex, blue] (s) -- (p)  node[pos=-0.2, left=1.5mm]  {\scalebox{0.9}{\obj{p2P}}};
  \draw[-latex, darkorange] (lh) -- (s)  node[pos=0.5, right=0.1mm]  {\scalebox{0.9}{\obj{g}}};
  \draw[-latex, darkorange] (rh) -- (p)  node[pos=0.13, right=0.1mm]  {\scalebox{0.9}{\obj{g}}};

  \draw[-latex, darkorange] (lh) -- (s)  node[pos=0.5, right=0.1mm]  {\scalebox{0.9}{\obj{g}}};
  \draw[-latex, darkorange] (rh) -- (p)  node[pos=0.13, right=0.1mm]  {\scalebox{0.9}{\obj{g}}};

\end{tikzpicture}}
            & \resizebox{!}{27mm}{\begin{tikzpicture}
  \node[rectangle, rounded corners, minimum width=9mm, minimum height=4mm, fill=amber!30] at (-0.90, -0.34) {};
  \node[rectangle, rounded corners, minimum width=6mm, minimum height=6mm, fill=amber!30] at ( 0.30, -1.00) {};
  
  \node[draw, fill=blue!15, ellipse, minimum width=8mm, minimum height=7mm] (vs) at ( 0.0,  0.0) {\obj{vs}};       
  \node[draw, fill=green!15, ellipse, minimum width=8mm, minimum height=7mm] (p) at (-1.0, -1.0) {\obj{pt}};       
  \node[draw, fill=green!15, ellipse, minimum width=8mm, minimum height=7mm] (s) at (-0.0, -2.0) {\obj{sp}};
  
  \node[draw, fill=red!15, ellipse, minimum width=8mm, minimum height=7mm] (lh) at ( 0.0, -3) {\obj{lh}};
  \node[draw, fill=red!15, ellipse, minimum width=8mm, minimum height=7mm] (rh) at (-1.0, -3) {\obj{rh}};

  \draw[-latex, blue] (p) -- (vs) node[pos=0.7, left]       {\scalebox{0.9}{2\obj{\times p2P}}};
  \draw[-latex, blue] (s) -- (vs) node[pos=0.6, right=0.1mm]  {\scalebox{0.9}{2\obj{\times}}};
  \draw[-latex, blue] (s) -- (vs) node[pos=0.4, right=0.1mm]  {\scalebox{0.9}{\obj{p2P}}};
  \draw[-latex, blue] (s) -- (p)  node[pos=0.3, left=0.5mm]   {\scalebox{0.9}{\obj{p2p}}};
  \draw[-latex, blue] (s) -- (p)  node[pos=-0.2, left=1mm]  {\scalebox{0.9}{\obj{p2P}}};
  \draw[-latex, darkorange] (lh) -- (s)  node[pos=0.5, right=0.1mm]  {\scalebox{0.9}{\obj{g}}};
  \draw[-latex, darkorange] (rh) -- (p)  node[pos=0.13, right=0.1mm]  {\scalebox{0.9}{\obj{g}}};
  
  \draw[-latex, darkorange] (lh) -- (s)  node[pos=0.5, right=0.1mm]  {\scalebox{0.9}{\obj{g}}};
  \draw[-latex, darkorange] (rh) -- (p)  node[pos=0.13, right=0.1mm]  {\scalebox{0.9}{\obj{g}}};
  
\end{tikzpicture}}
            & \resizebox{!}{27mm}{\begin{tikzpicture}
  \node[draw, fill=blue!15, ellipse, minimum width=8mm, minimum height=7mm] (p) at ( 0.0,  0.0) {\obj{pt}};       
  \node[draw, fill=green!15, ellipse, minimum width=8mm, minimum height=7mm] (s) at (-0.0, -2.0) {\obj{sp}};
  
  \node[draw, fill=red!15, ellipse, minimum width=8mm, minimum height=7mm] (lh) at ( 0.0, -3) {\obj{lh}};

  \draw[-latex, blue] (s) -- (p) node[pos=0.7, right=0.1mm]  {\scalebox{0.9}{\obj{p2p}}};
  \draw[-latex, blue] (s) -- (p) node[pos=0.4, right=0.1mm]  {\scalebox{0.9}{2\obj{\times}}};
  \draw[-latex, blue] (s) -- (p) node[pos=0.2, right=0.1mm]  {\scalebox{0.9}{\obj{p2P}}};
  \draw[-latex, darkorange] (lh) -- (s) node[pos=0.5, right=0.1mm]  {\scalebox{0.9}{\obj{g}}};
  \draw[-latex, darkorange] (lh) -- (s) node[pos=0.5, right=0.1mm]  {\scalebox{0.9}{\obj{g}}};
  
\end{tikzpicture}}
        \\
        \bottomrule
    \end{tabular}
    \caption{
        Loosely-coupled task: place spoon (\taskabbr{Pl_{sp}}) on a plate with different styles. Objects include a spoon (\obj{sp}), a plate (\obj{pt}), and two hands (\obj{lh,rh}). We prefix the letter \obj{v} to the corresponding virtual object, \eg \obj{vs} stands for the virtual spoon. 
    }
    \label{table:eval_psp}
    \vspace{-2.5ex}
\end{table*}


\subsection{Bimanual Coordination Strategies}\label{subsec:coordination}
Given the grasp relationship and HMSR extracted in \cref{subsec:hms}, we derive bimanual coordination strategies.

\subsubsection{Uncoordinated unimanual} A single grasp relationship is detected between a hand and an object. This corresponds to the K-VIL, \ie, single MSR case (see also \cref{subfig:hmsr_single}). 

\subsubsection{Uncoordinated bimanual} Each hand grasps a different \slave{} object, and these two \slave{} objects have different \master{} objects (see \cref{table:eval_other}). 
In this case, the two hands perform different tasks without coordination.


\subsubsection{Loosely-coupled coordination} Interaction forces between two hand groups, \ie, the union of the hand and a grasped object, is key to distinguishing loosely-coupled and tightly-coupled asymmetric coordination strategies~\cite{krebs_bimanual_2022}. 
Since estimating interaction forces from visual demonstrations is not trivial, we do not distinguish between these two strategies and group them into a single one. 
That is, if constraints exist between the objects grasped individually by each hand, or if both grasped \slave{} objects share at least one \master{} object, the two hand groups are loosely-coupled. In the former case, one hand is constrained by another, whereas in the latter, the two hands move toward the same \master{} object.


\subsubsection{Tightly-coupled symmetric coordination} For a noticeable time window,  
\begin{enumerate*}[label=(\roman*)]
    \item both hands grasp the same object, and 
    \item the distance change rate (see \cref{subsec:rel_saliency}) between two hands drops below a certain threshold.
\end{enumerate*}


\section{Bimanual Keypoint-based Admittance Controller (Bi-KAC)}
\label{sec:bikac}

Given the subsymbolic task representations in the HMSR graph including the keypoints, their associated local frames, geometric constraints and MPs,
we derive a compliant and torque-controlled bimanual keypoint-based controller extended from KAC~\cite{gao_kvil_2023}.
We control the Tool-Center-Point (TCP) of each robot hand with an impedance controller. 
The TCP task space target is derived using KAC for each arm.
Specifically, the spring-damper systems of all constraints defined for an object in a hand contribute to the forces driving this hand.
The coordination is achieved via the HMSR. In other words, Bi-KAC is a naive extension of KAC, which handles bimanual coordination via the HMSR representation.
For example, in \cref{subfig:teaser_exe}, the left hand grasps the kettle following $\ptop$ and $\ptoc$ constraints. 
The corresponding VMPs and constraints are dynamically updated by the moving \master{} cup grasped by the right hand, which itself is controlled by a task-space VMP towards a pose constraint defined on the static virtual cup.


\section{Evaluation}
\label{sec:eval}

We evaluate our approach in eight real-world tasks, namely, 
pour water (\taskabbr{Po_w}), pour beer (\taskabbr{Po_b}), place spoon (\taskabbr{Pl_{sp}}), place serving tray (\taskabbr{Pl_{st}}), place spoon and plate (\taskabbr{Pl_{sp,pt}}), place cutboard and pan (\taskabbr{Pl_{cb,pa}}), place spoon and banana (\taskabbr{Pl_{sp,ba}}), and clean table (\taskabbr{C_{ta}}).
Given a few demonstration videos of each task recorded with Azure Kinect or Stereolab ZED camera, we run our perception pipeline to obtain the 3D point trajectories of objects and hands, extract a HMSR and a coordination strategy, and reproduce the tasks with Bi-KAC in novel scenes.
We evaluate Bi-KVIL's ability to 
\begin{enumerate*}
    \item extract a consistent HMSR from different styles of task demonstrations,
    \item capture these fine-grained styles in its sub-symbolic task representation, and
    \item reproduce the learned tasks with categorical generalization. 
\end{enumerate*}


\subsection{Task Extraction}\label{subsec:task_extr}
For each task, we provide different styles and numbers \scalebox{0.9}{\circled{N}} of demonstrations, resulting in a total of 14 evaluations. 
Specifically, in the \taskabbr{Pl_{sp}} task, the motion styles are:  
\begin{enumerate*}
    \item[(\taskabbr{Pl^{1}_{sp}})] the plate moves to the spoon and the spoon is lifted up and placed at the center of the plate,
    \item[(\taskabbr{Pl^{2}_{sp}})] as \taskabbr{Pl^{1}_{sp}} but plates are taken from various positions above the table,
    \item[(\taskabbr{Pl^{3}_{sp}})] similar to \taskabbr{Pl^{1}_{sp}}, but the spoon is placed at an arbitrary position on the plate,
    \item[(\taskabbr{Pl^{4}_{sp}})] the plate moves to an arbitrary position with a spoon at the center, and
    \item[(\taskabbr{Pl^{5}_{sp}})] unimanual placement.
\end{enumerate*}
Results are displayed in~\cref{table:eval_psp}). 
With \circled{3} and \circled{4} demonstrations in \taskabbr{Pl^{1}_{sp}} with small pose variations of plates with respect to the virtual plate, Bi-KVIL extracts more $\ptop$ (\raisebox{-1.3mm}{\sampleline{gray!20, line width=3mm}}) constraints than for the other tasks.
With additional demonstrations in \task{Pl^{1}_{sp}}{5}/\scalebox{0.9}{\circled{6}}, $\ptoP$ constraints for the spoon are extracted with respect to multiple \master{} objects, \ie, the plate, virtual spoon, and virtual plate. 
Since the plate always remains on the table surface, it is reasonable that multiple $\ptoP$ constraints exist.
When the plate starts from above the table in \task{Pl^{2}_{sp}}{6}, Bi-KVIL learns to eliminate the redundant \master{}-\slave{} pairs related to the virtual plate and the associated $\ptoP$ constraints (\sampleline{dashed, red, line width=1.5pt, dash pattern=on 2pt off 2pt}), resulting in a more compact HMSR graph.
Compared to \task{Pl^{2}_{sp}}{6}, the spoons are placed at arbitrary positions on the plate in \task{Pl^{3}_{sp}}{6}, so that the $\ptop$ constraints between the spoon and its two \master{}s (\raisebox{-1.3mm}{\sampleline{amber!30, line width=3mm}}) are truncated and an additional $\ptoP$ constraint is created.
Similarly, in \task{Pl^{4}_{sp}}{6}, the plate is constrained only by the table surface and the spoon follows its motion, so that $\ptop$ (\raisebox{-1.3mm}{\sampleline{amber!30, line width=3mm}}) are replaced by $\ptoP$ constraints.
Except for the redundant \master{}-\slave{} pair of \task{Pl^{1}_{sp}}{5}/\scalebox{0.9}{\circled{6}} and the unimanual case \task{Pl^{5}_{sp}}{6}, the HMSR graph is structured identically across task styles, but differs in sub-symbolic constraints as different motion styles are captured. Moreover, redundant relations and constraints are eliminated by providing more demonstrations with variations. 
For all bimanual \taskabbr{Pl_{sp}} tasks, Bi-KVIL extracts a loosely-coupled bimanual coordination strategy with the right hand group being non-dominant since the plate is a \master{} of the spoon.

\begin{table}
    \centering
    \setlength\tabcolsep{0.5pt}
    \setlength\imgwidth{19mm}
    \begin{tabular}{cM{\imgwidth}M{\imgwidth}M{\imgwidth}M{\imgwidth}}
       \toprule
            Tasks 
            & \task{Pl_{st}}{6} 
            & \task{Pl_{cb,pa}}{8}
            & \task{Pl_{sp,pt}}{6}
            & \task{Pl_{sp,ba}}{6}
        \\
        \midrule
            \multirow{1}{*}[2em]{\rotatebox[origin=c]{90}{Demo.}}
            & \multicolumn{1}{c}{\includegraphics[height=10mm, trim=300 80 220 220, clip]{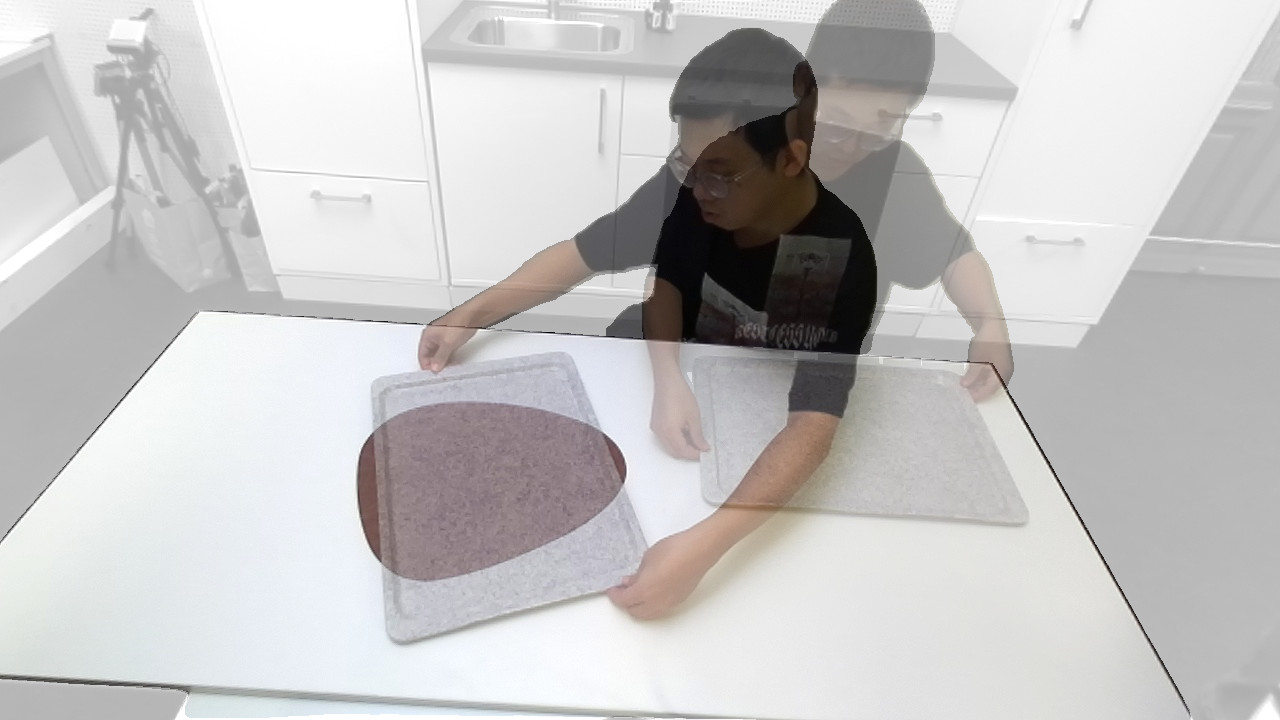}}
            & \multicolumn{1}{c}{\includegraphics[height=10mm, trim=300 100 300 200, clip]{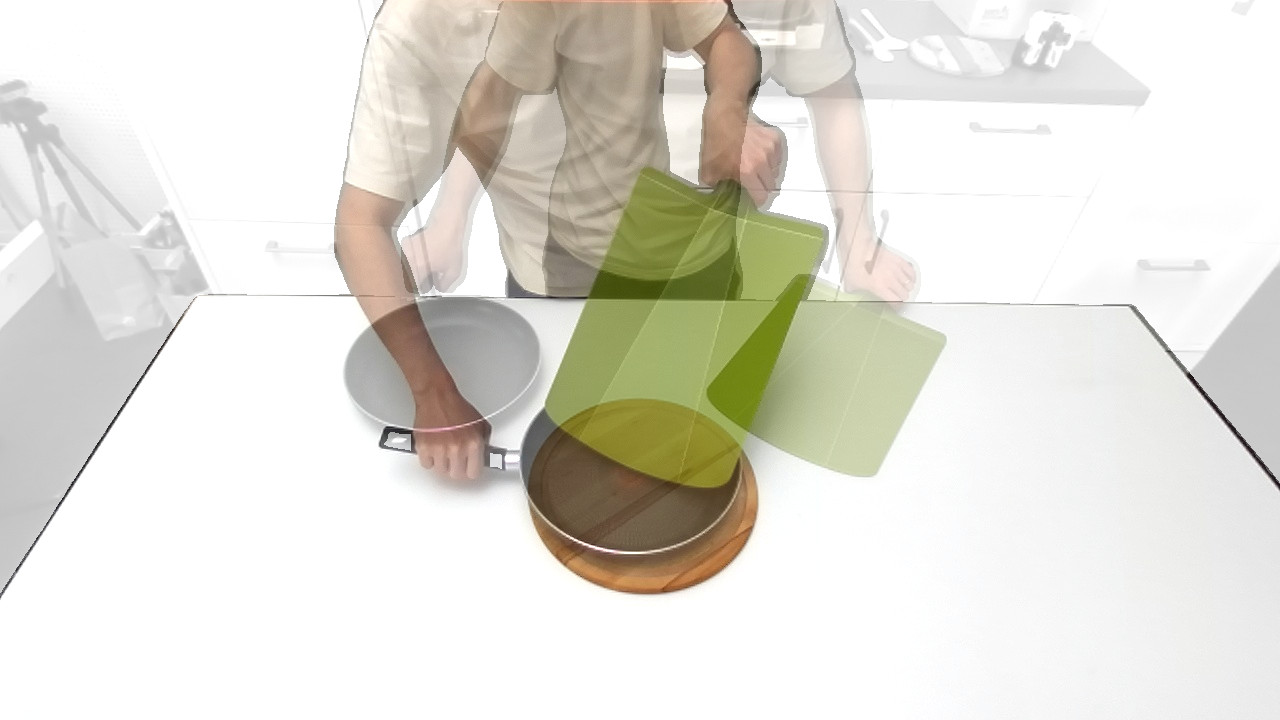}}
            & \multicolumn{1}{c}{\includegraphics[height=10mm, trim=220 50 150 220, clip]{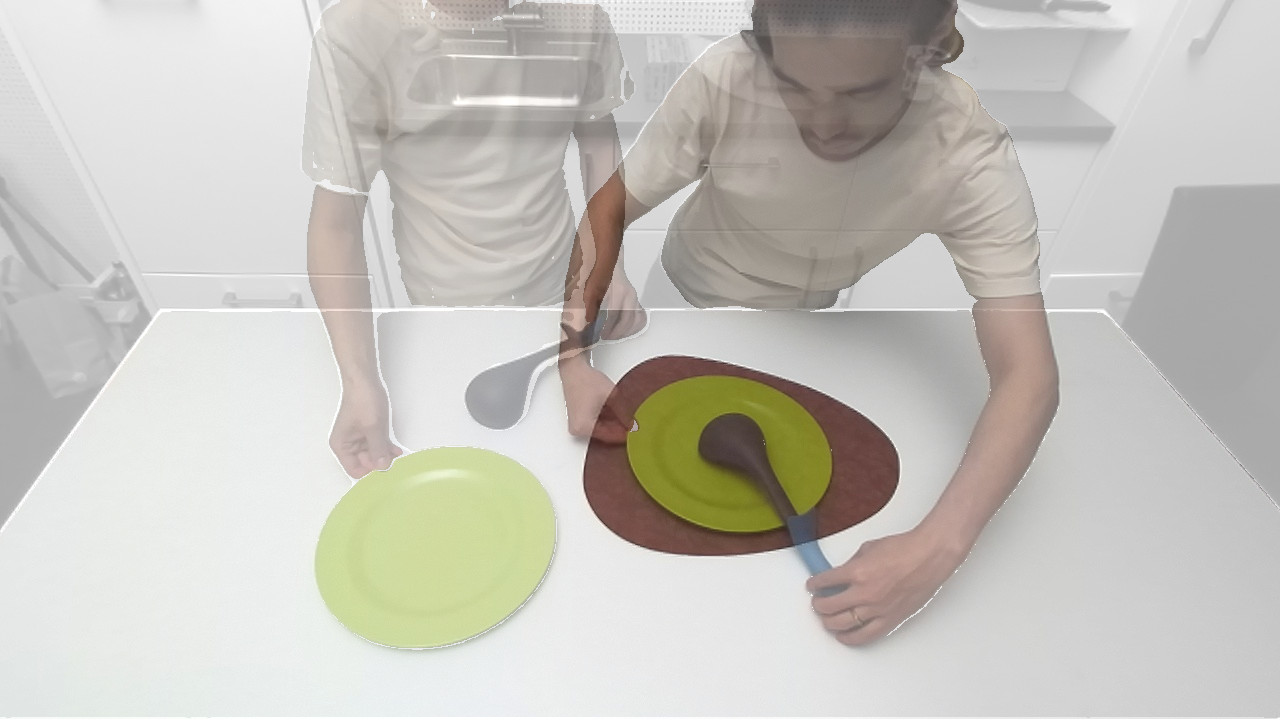}}
            & \multicolumn{1}{c}{\includegraphics[height=10mm, trim=250 20 150 200, clip]{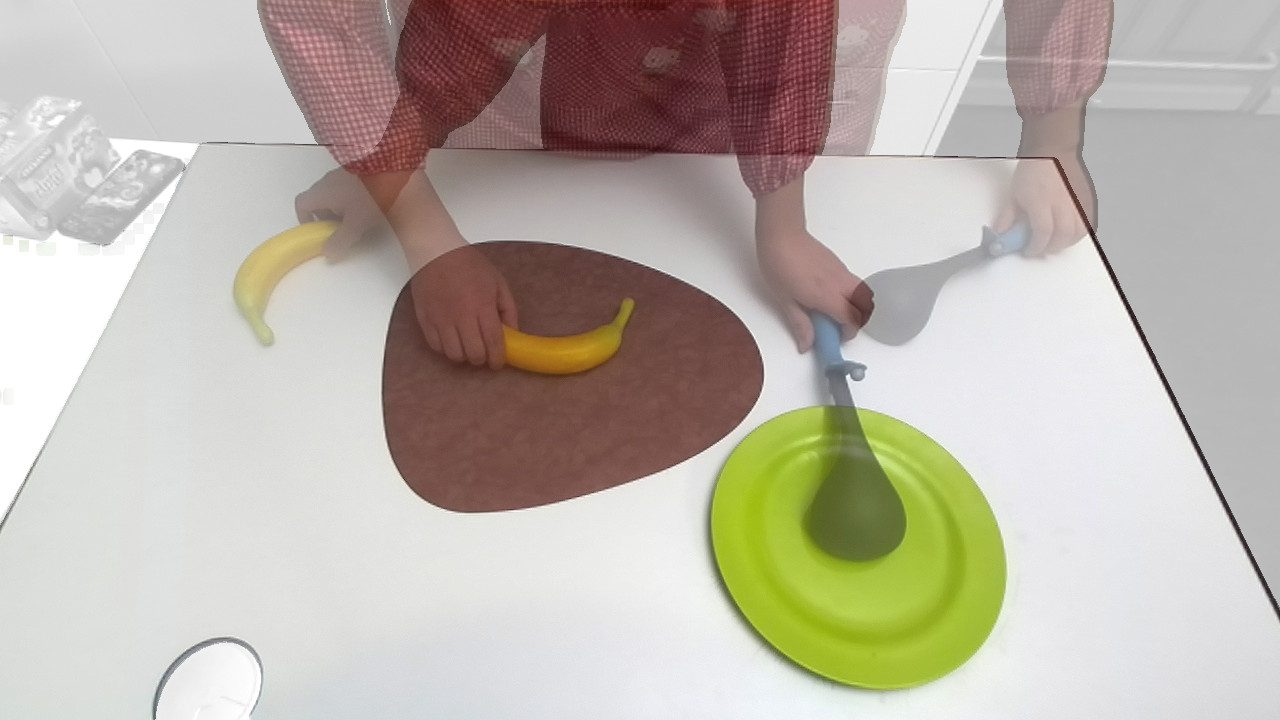}}
        \\
            \multirow{1}{*}[1em]{\rotatebox[origin=c]{90}{HMSR}}
            & \resizebox{!}{27mm}{\begin{tikzpicture}
  
  \node[draw, fill=blue!15,  ellipse, minimum width=8mm, minimum height=7mm] (t) at ( 0.0,  0.0) {\obj{tm}};       
  \node[draw, fill=green!15, ellipse, minimum width=8mm, minimum height=7mm] (s)  at (-0.0, -2.0) {\obj{st}};
  
  \node[draw, fill=red!15,   ellipse, minimum width=8mm, minimum height=7mm] (lh) at ( 0.7, -3.0) {\obj{lh}};
  \node[draw, fill=red!15,   ellipse, minimum width=8mm, minimum height=7mm] (rh) at (-0.7, -3.0) {\obj{rh}};

  \draw[-latex, blue] (s) -- (t)  node[pos=0.7, left]         {\scalebox{0.9}{\obj{p2p}}};
  \draw[-latex, blue] (s) -- (t)  node[pos=0.4, left]         {\scalebox{0.9}{2\obj{\times p2P}}};
  \draw[-latex, darkorange] (lh) -- (s)  node[pos=0.5, right=0.1mm]  {\scalebox{0.9}{\obj{g}}};
  \draw[-latex, darkorange] (rh) -- (s)  node[pos=0.5, left=0.1mm]   {\scalebox{0.9}{\obj{g}}};
  \draw[-latex, darkorange] (lh) -- (s)  node[pos=0.5, right=0.1mm]  {\scalebox{0.9}{\obj{g}}};
  \draw[-latex, darkorange] (rh) -- (s)  node[pos=0.5, left=0.1mm]   {\scalebox{0.9}{\obj{g}}};

  \draw[double, darkpink, double distance=2pt] (lh) -- (rh) node[pos=0.5, above] {\scalebox{0.9}{\obj{sym}}};
  \draw[double, darkpink, double distance=2pt] (lh) -- (rh) node[pos=0.5, above] {\scalebox{0.9}{\obj{sym}}};
  
\end{tikzpicture}}
            & \resizebox{!}{27mm}{\begin{tikzpicture}
  
  \node[draw, fill=blue!15, ellipse, minimum width=8mm, minimum height=7mm]  (m) at (-1.0,  0.0) {\obj{pm}};       
  \node[draw, fill=green!15, ellipse, minimum width=8mm, minimum height=7mm] (p) at (-1.0, -1.3) {\obj{pa}};       
  \node[draw, fill=green!15, ellipse, minimum width=8mm, minimum height=7mm] (c) at (-0.0, -2.0) {\obj{cb}};
  
  \node[draw, fill=red!15, ellipse, minimum width=8mm, minimum height=7mm] (lh) at ( 0.0, -3) {\obj{lh}};
  \node[draw, fill=red!15, ellipse, minimum width=8mm, minimum height=7mm] (rh) at (-1.0, -3) {\obj{rh}};

  \draw[-latex, blue] (p) -- (m)  node[pos=0.8, right]      {\scalebox{0.9}{\obj{p2p, p2c}}};
  \draw[-latex, blue] (p) -- (m)  node[pos=0.2, right]      {\scalebox{0.9}{2\obj{\times p2P}}};
  
  \draw[-latex, blue] (c) -- (p) node[pos=1.0, right=1.0mm] {\scalebox{0.9}{\obj{p2p, p2P}}};
  
  \draw[-latex, orange] (lh) -- (c)  node[pos=0.5, right=0.1mm]  {\scalebox{0.9}{\obj{g}}};
  \draw[-latex, orange] (rh) -- (p)  node[pos=0.13, right=0.1mm]  {\scalebox{0.9}{\obj{g}}};
  \draw[-latex, orange] (lh) -- (c)  node[pos=0.5, right=0.1mm]  {\scalebox{0.9}{\obj{g}}};
  \draw[-latex, orange] (rh) -- (p)  node[pos=0.13, right=0.1mm]  {\scalebox{0.9}{\obj{g}}};

\end{tikzpicture}}
            & \resizebox{!}{27mm}{\begin{tikzpicture}
  
  \node[draw, fill=blue!15, ellipse, minimum width=8mm, minimum height=7mm]  (t) at ( 0.0,  0.0) {\obj{tm}};       
  \node[draw, fill=green!15, ellipse, minimum width=8mm, minimum height=7mm] (p) at (-1.0, -1.0) {\obj{pt}};    
  \node[draw, fill=green!15, ellipse, minimum width=8mm, minimum height=7mm] (c) at (-0.0, -2.0) {\obj{sp}};
  
  \node[draw, fill=red!15, ellipse, minimum width=8mm, minimum height=7mm] (lh) at ( 0.0, -3) {\obj{lh}};
  \node[draw, fill=red!15, ellipse, minimum width=8mm, minimum height=7mm] (rh) at (-1.0, -3) {\obj{rh}};

  \draw[-latex, blue] (p) -- (t)  node[pos=0.8, left]      {\scalebox{0.9}{4\obj{\times p2p}}};
  \draw[-latex, blue] (s) -- (t)  node[pos=0.6, right]      {\scalebox{0.9}{\obj{p2p}}};
  \draw[-latex, blue] (s) -- (t)  node[pos=0.4, right]      {\scalebox{0.9}{\obj{p2P}}};
  
  \draw[-latex, blue] (s) -- (p) node[pos=0.3, left=0.1mm] {\scalebox{0.9}{\obj{p2p}}};
  \draw[-latex, blue] (s) -- (p) node[pos=-0.2, left=1mm] {\scalebox{0.9}{\obj{p2P}}};
  
  \draw[-latex, darkorange] (lh) -- (s)  node[pos=0.5, right=0.1mm]  {\scalebox{0.9}{\obj{g}}};
  \draw[-latex, darkorange] (rh) -- (p)  node[pos=0.13, right=0.1mm]  {\scalebox{0.9}{\obj{g}}};
  \draw[-latex, darkorange] (lh) -- (s)  node[pos=0.5, right=0.1mm]  {\scalebox{0.9}{\obj{g}}};
  \draw[-latex, darkorange] (rh) -- (p)  node[pos=0.13, right=0.1mm]  {\scalebox{0.9}{\obj{g}}};

\end{tikzpicture}}
            & \resizebox{!}{27mm}{\begin{tikzpicture}
  
  \node[draw, fill=blue!15, ellipse, minimum width=8mm, minimum height=7mm]  (t) at (-1.0,  0.0) {\obj{tm}};    
  \node[draw, fill=blue!15, ellipse, minimum width=8mm, minimum height=7mm]  (p) at ( 0.0,  0.0) {\obj{pt}};  
  \node[draw, fill=green!15, ellipse, minimum width=8mm, minimum height=7mm] (b) at (-1.0, -2.0) {\obj{ba}};       
  \node[draw, fill=green!15, ellipse, minimum width=8mm, minimum height=7mm] (s) at (-0.0, -2.0) {\obj{sp}};
  
  \node[draw, fill=red!15, ellipse, minimum width=8mm, minimum height=7mm] (lh) at ( 0.0, -3) {\obj{lh}};
  \node[draw, fill=red!15, ellipse, minimum width=8mm, minimum height=7mm] (rh) at (-1.0, -3) {\obj{rh}};

  \draw[-latex, blue] (b) -- (t)  node[pos=0.7, right=0.1mm]  {\scalebox{0.9}{\obj{p2p}}};
  \draw[-latex, blue] (b) -- (t)  node[pos=0.4, right=0.1mm]  {\scalebox{0.9}{2\obj{\times}}};
  \draw[-latex, blue] (b) -- (t)  node[pos=0.2, right=0.1mm]  {\scalebox{0.9}{\obj{\ptoP}}};

  \draw[-latex, blue] (s) -- (p)  node[pos=0.7, right=0.1mm]  {\scalebox{0.9}{\obj{p2p}}};
  \draw[-latex, blue] (s) -- (p)  node[pos=0.4, right=0.1mm]  {\scalebox{0.9}{2\obj{\times}}};
  \draw[-latex, blue] (s) -- (p)  node[pos=0.2, right=0.1mm]  {\scalebox{0.9}{\obj{p2P}}};
  
  \draw[-latex, darkorange] (lh) -- (s)  node[pos=0.5, right=0.1mm]  {\scalebox{0.9}{\obj{g}}};
  \draw[-latex, darkorange] (rh) -- (b)  node[pos=0.5, right=0.1mm]  {\scalebox{0.9}{\obj{g}}};
  \draw[-latex, darkorange] (lh) -- (s)  node[pos=0.5, right=0.1mm]  {\scalebox{0.9}{\obj{g}}};
  \draw[-latex, darkorange] (rh) -- (b)  node[pos=0.5, right=0.1mm]  {\scalebox{0.9}{\obj{g}}};

\end{tikzpicture}}
        \\
        \bottomrule
    \end{tabular}
    \caption{
        HMSR for \taskabbr{Pl_{st}}, \taskabbr{Pl_{cb,pa}}, \taskabbr{Pl_{sp,pt}}, and \taskabbr{Pl_{sp,ba}} tasks. 
        Legend as \cref{table:eval_psp} and 
        \obj{tm, sp, pm, cb, pt, ba} stand for tablemat, spoon, potmat, cutboard, plate, and banana.
    }
    \label{table:eval_other}
    \vspace{-1.5ex}
\end{table}

\begin{table}
    \centering
    \setlength\tabcolsep{0.5pt}
    \setlength\imgwidth{19mm}
    \begin{tabular}{cM{\imgwidth}M{\imgwidth}M{\imgwidth}M{\imgwidth}}
       \toprule
            Task
            & RGB
            & TR
            & Start
            & End
        \\
        \midrule
            \multirow{1}{*}[1em]{\rotatebox[origin=c]{90}{\task{Pl^{2}_{sp}}{6}}}
            & 
            \resizebox{!}{10mm}{
                \includegraphics[height=10mm]{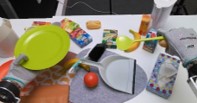}
            }
            & 
            \resizebox{!}{10mm}{
                \input{figure/eval/exe_tr/plate_to_spoon_tr.tikz}
            }
            & 
            \resizebox{!}{10mm}{
                \includegraphics[height=10mm]{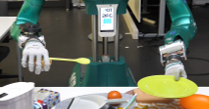}
            }
            & \resizebox{!}{10mm}{
                \includegraphics[height=10mm]{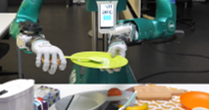}
            }
        \\ 
            \multirow{1}{*}[1em]{\rotatebox[origin=c]{90}{\task{Pl^{3}_{sp}}{6}}}
            & 
            \resizebox{!}{10mm}{
                \includegraphics[height=10mm]{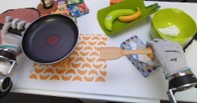}
            }
            & \resizebox{!}{10mm}{
                \input{figure/eval/exe_tr/spoon_to_plane_tr.tikz}
            }
            & 
            \resizebox{!}{10mm}{
                \includegraphics[height=10mm]{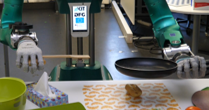}
            }
            & 
            \resizebox{!}{10mm}{
                \includegraphics[height=10mm]{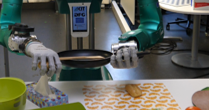}
            }
        \\ 
            \multirow{1}{*}[12pt]{\rotatebox[origin=c]{90}{
                \task{Pl^{4}_{sp}}{6}
            }}
            & 
            \resizebox{!}{10mm}{
                \includegraphics[height=10mm]{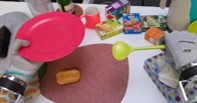}
            }
            & 
            \resizebox{!}{10mm}{
            \input{figure/eval/exe_tr/plate_to_any_tr.tikz}
            }
            & 
            \resizebox{!}{10mm}{
                \includegraphics[height=10mm]{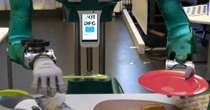}
            }
            & \resizebox{!}{10mm}{
                \includegraphics[height=10mm]{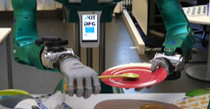}
            }
        \\ 
            \multirow{1}{*}[12pt]{\rotatebox[origin=c]{90}{
                \task{Pl^{5}_{sp}}{6}
            }}
            & \resizebox{!}{10mm}{
                \includegraphics[height=10mm]{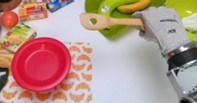}
            }
            & \resizebox{!}{10mm}{
                \input{figure/eval/exe_tr/uni_tr.tikz}
            }
            & \resizebox{!}{10mm}{
                \includegraphics[height=10mm]{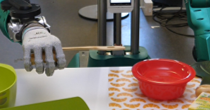}
            }
            & \resizebox{!}{10mm}{
                \includegraphics[height=10mm]{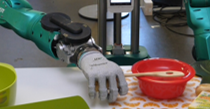}
            }
        \\
        \bottomrule
    \end{tabular}
    \caption{
        Reproduction of the \taskabbr{Pl_{sp}} tasks with different styles corresponding to \task{Pl^{2-5}_{sp}}{6}. Images in each row correspond to RGB perception from the robot's viewpoint, the task representation (TR), and the start and end of execution.} 
    \label{table:eval_exe}
    \vspace{-1.5ex}
\end{table}

As shown in \cref{fig:overview}, different pouring styles also share the same HMSR structure at a symbolic level and differ only in the sub-symbolic definition of the $\ptop$, $\ptoc$, and pose constraints. 
Note that the pose constraints, required to tilt the cup in the \taskabbr{Po_{b}} task, are correctly modeled and reproduced by Bi-KAC (see \cref{subfig:teaser_constr,subfig:teaser_exe}). 
In \cref{subfig:teaser_constr}, the cup travels longer and faster on average than the kettle or beer bottle. 
In contrast to the rule-based algorithm in~\cite{krebs_bimanual_2022}, our pose invariance criteria identify the cup as \master{} as it displays less orientation variation.

The virtual object is not required in the presence of a static real object from which the spatial invariances of the moving objects are more salient. 
When the demonstrations contain sufficient pose or shape variations, Bi-KVIL truncates the virtual objects in the \taskabbr{Pl_{st}}, \taskabbr{Pl_{cb,pa}}, \taskabbr{Pl_{sp,pt}}, and \taskabbr{Pl_{sp,ba}} tasks, and a static real object serves as the top-level \master{} object (see~\cref{table:eval_other}).
In \task{Pl_{st}}{6}, symmetric coordination is extracted along with sub-symbolic constraints $\ptop$ and $\ptoP$ defining the target pose of the serving tray right above the center of the tablemat. 
Bi-KVIL also deals with tasks involving more than two objects, \eg, \task{Pl_{cb,pa}}{8}, \task{Pl_{sp,pt}}{6} and \task{Pl_{sp,ba}}{6}, where a loosely-coupled coordination is extracted for the former two and uncoordinated bimanual coordination for the latter. 
The \master{}-\slave{} pairs between the hand groups are truncated as K-VIL finds no salient geometric constraint.

\subsection{Task Reproduction}\label{subsec:bikac_reproduction}
We evaluate Bi-KAC qualitatively for each task in \cref{subsec:task_extr} and refer the reader to~\cite{gao_kvil_2023} for quantitative evaluations, as its behavior for each arm inherits KAC.
Here, we select one example per style of the \taskabbr{Pl_{sp}} task to illustrate the behavior of Bi-KAC in reproducing the learned task with the \armarVI humanoid robot~\cite{Asfour2019}.
As shown in \cref{table:eval_exe}, the plate is driven to the initial position of the spoon with the spoon head right above the center of the plate in \task{Pl^{2}_{sp}}{6}. 
This is due to the $\ptop$ constraints between the plate and the virtual spoon and between the spoon and the plate.
In contrast, the plate in \task{Pl^{4}_{sp}}{6} moves on plane constraints, and the spoon is placed anywhere on the pan in \task{Pl^{3}_{sp}}{6} as $\ptop$ constraints were eliminated (\raisebox{-1.3mm}{\sampleline{amber!30, line width=3mm}}).
Notice that all tasks were also reproduced using out-of-distribution objects such as spoons of various shapes, plates of different sizes and colors, and cooking pans instead of plates in \taskabbr{Pl_{sp}} in \cref{table:eval_exe}.
We refer the interested reader to our website \href{https://sites.google.com/view/bi-kvil}{https://sites.google.com/view/bi-kvil} for results of other tasks with different styles, numbers of demonstrations, and out-of-distribution objects.


\section{Conclusion}
\label{sec:conclusion}

In this paper, we proposed Bi-KVIL, a novel keypoints-based approach for visual imitation learning of bimanual manipulation tasks. Bi-KVIL simultaneously extracts hybrid master-slave relationships (HMSR) and bimanual coordination strategies at the symbolic level, as well as the task representations capturing the fine-grained motion styles at the sub-symbolic level. 
The proposed HMSR covers the bimanual manipulation taxonomy~\cite{krebs_bimanual_2022} and enables unified keypoints-based bimanual controllers for both uni- and bimanual tasks. 
By explicitly modeling the master-slave relationships and geometric constraints in an object-centric manner, our representation is embodiment-independent and viewpoint invariant (see~\cite[Section VII]{gao_kvil_2023}), and generalizes well to categorical objects. 
Bi-KVIL allows us to learn bimanual task representations while requiring less than 10 human demonstration videos from RGB-D cameras without additional devices.
In comparison, other bimanual imitation learning approaches demand a large number of demonstrations, \eg, 20 to 50 in~\cite{zhao_learning_2023,mobile_aloha}, 2500 to 4700 in~\cite{xie_deep_2020}, and 256 to 4000 in~\cite{kim_robot_2022a,kim_transformerbased_2021}. 
Some approaches additionally require teleoperation data~\cite{zhao_learning_2023,mobile_aloha} or human pose recorded using motion capture system~\cite{liu_robot_2022}.
Finally, we establish a perception pipeline leveraging advanced computer vision algorithms to provide high-quality datasets for VIL.

Although our perception pipeline includes hand shape completion~\cite{lin_mesh_2021}, it does not handle object (self-)occlusion, leading to failures if keypoints are occluded.
Failures may also occur due to inacurate correspondence detection in specific objects poses, which lead to imprecise local frames on master objects and 
inaccurate target positions, \eg, the spout of the kettle being outside the cup rim. 
Moreover, Bi-KAC, as a naive extension of KAC, relies entirely on the HMSR for coordination and disregards dual-arm synchronization~\cite{gao_projected_2018,lin_projected_2018,shahriari_passivity_2022}. Therefore, it may drop the object in bimanual transport tasks.
In future work, we plan to address these limitations and to investigate a comprehensive evaluation benchmark for bimanual imitation learning tasks.


\bibliographystyle{IEEEtran}
\bibliography{jianfeng_library_simplified,HumanoidsGroup}


\clearpage
\end{document}